\pdfoutput=1

\documentclass[11pt]{article}

\usepackage[final]{acl}

\usepackage{times}
\usepackage{latexsym}

\usepackage[T1]{fontenc}

\usepackage[utf8]{inputenc}

\usepackage{microtype}

\usepackage{inconsolata}

\usepackage{graphicx}

\usepackage{amsmath}
\usepackage{amssymb}
\usepackage{multirow}
\usepackage{multicol}
\usepackage{graphicx}
\usepackage{enumitem}
\usepackage{hyperref}
\usepackage{tikz}
\newcommand*{\circled}[1]{\lower.7ex\hbox{\tikz\draw (0pt, 0pt)%
    circle (.5em) node {\makebox[1em][c]{\scriptsize #1}};}}

\title{Multilingual Machine Translation with Open Large Language Models at Practical Scale: An Empirical Study}
\author{Menglong Cui{\thanks{~Equal contribution. Correspondence to: Pengzhi Gao <gaopengzhi@xiaomi.com>.}}, Pengzhi Gao{\footnotemark[1]}, Wei Liu, Jian Luan, Bin Wang \\
Xiaomi Inc., Beijing, China \\
\texttt{cuimenglongcs@gmail.com} \\
\texttt{\{gaopengzhi,liuwei40,luanjian,wangbin11\}@xiaomi.com}
}

\begin{document}

\maketitle

\begin{abstract}

Large language models (LLMs) have shown continuously improving multilingual capabilities, and even small-scale open-source models have demonstrated rapid performance enhancement. In this paper, we systematically explore the abilities of open LLMs with less than ten billion parameters to handle multilingual machine translation (MT) tasks. We conduct comprehensive evaluations on six popular LLMs and find that models like Gemma2-9B exhibit impressive multilingual translation capabilities. We then introduce the \textbf{P}arallel-\textbf{F}irst \textbf{M}onolingual-\textbf{S}econd (PFMS) data mixing strategy in the continual pretraining stage to further enhance the MT performance and present GemmaX2-28, a 9B model achieving top-tier multilingual translation performance across 28 languages\footnote{Arabic (\texttt{ar}), Bengali (\texttt{bn}), Czech (\texttt{cs}), German (\texttt{de}), English (\texttt{en}), Spanish (\texttt{es}), Persian (\texttt{fa}), French (\texttt{fr}), Hebrew (\texttt{he}), Hindi (\texttt{hi}), Indonesian (\texttt{id}), Italian (\texttt{it}), Japanese (\texttt{ja}), Khmer (\texttt{km}), Korean (\texttt{ko}), Lao (\texttt{lo}), Malay (\texttt{ms}), Burmese (\texttt{my}), Dutch (\texttt{nl}), Polish (\texttt{pl}), Portuguese (\texttt{pt}), Russian (\texttt{ru}), Thai (\texttt{th}), Tagalog (\texttt{tl}), Turkish (\texttt{tr}), Urdu (\texttt{ur}), Vietnamese (\texttt{vi}), and Chinese (\texttt{zh}).}. Specifically, GemmaX2-28 consistently outperforms the state-of-the-art (SOTA) models such as TowerInstruct \cite{alves2024tower} and X-ALMA \cite{xu2024xalmaplugplay} and achieves competitive performance with Google Translate and GPT-4-turbo. \footnote{Models are released at \href{https://huggingface.co/collections/ModelSpace/gemmax2-673714f5049bfa3a90bee6b6}{https://huggingface/GemmaX2}.}

\end{abstract}

\section{Introduction}

Large language models (LLMs) such as GPT models \cite{openai2023chatgpt,openai2023gpt4}, PaLM models \cite{chowdhery2022palm,anil2023palm}, and others have demonstrated remarkable capabilities in multilingual translation \cite{jiao2023chatgpt,vilar-etal-2023-prompting}. Recently, significant efforts have been directed towards bolstering the multilingual competencies of open LLMs across a diverse linguistic spectrum, beyond English and Chinese. For example, over 5\% of the LLaMA3 \cite{dubey2024llama3herdmodels} pretraining datasets consist of high-quality non-English data covering over $30$ languages; Qwen2/2.5 \cite{yang2024qwen2technicalreport} claim to have multilingual support for over $29$ languages; \citet{ustun-etal-2024-aya} introduce Aya-101, a massively multilingual generative language model supporting $101$ languages.

\citet{zhu-etal-2024-multilingual} evaluate popular LLMs on multilingual MT tasks and find that the translation capability of LLMs is continually evolving. However, the models assessed (XGLM \cite{lin-etal-2022-shot}, OPT \cite{zhang2022optopenpretrainedtransformer}, Falcon \cite{almazrouei2023falconseriesopenlanguage}, BLOOMZ \cite{workshop2023bloom176bparameteropenaccessmultilingual}, LLaMA2 \cite{touvron2023llama2openfoundation}) are outdated due to the rapid development of open LLMs, and the multilingual translation ability of the latest models remains unclear. One natural question arises: 1) \textit{How do the latest open LLMs with practical scale perform multilingual MT tasks?}

To further boost LLMs' translation capability, existing approaches usually adopt multilingual corpora during continual pretraining on open LLMs. Specifically, \citet{DBLP:conf/iclr/Xu0SA24} utilize continual pretraining on monolingual datasets. \citet{guo-etal-2024-novel} apply continual pretraining on monolingual datasets followed by parallel datasets. \citet{alves2024tower} perform continual pretraining on a multilingual mixture of monolingual (two-thirds) and parallel (one-third) datasets. However, the optimal approach to leverage monolingual and parallel datasets for multilingual MT remains under-explored. Another natural question arises: 2) \textit{What are the best practices for leveraging monolingual and parallel corpora to enhance the multilingual MT performance of LLMs?}

In this paper, we first evaluate the in-context capabilities of the latest open-source LLMs including Mistral-7B-v0.3 \cite{jiang2023mistral7b}, Qwen2/2.5-7B \cite{yang2024qwen2technicalreport}, LLaMA3/3.1-8B \cite{dubey2024llama3herdmodels}, and Gemma2-9B \cite{gemmateam2024gemma2improvingopen} across $28$ languages on the multilingual MT tasks. We find that open LLMs like Gemma2-9B exhibit remarkable multilingual translation capabilities but still fall behind strong closed-source models. We then systematically investigate the most effective methods to leverage monolingual and parallel data for further boosting Gemma2-9B's capabilities and propose the Parallel-First Monolingual-Second (PFMS) data mixing strategy. Leveraging PFMS strategy when continually pretraining on Gemma2-9B followed by instruction finetuning on a small set of high-quality translation pairs, we learn GemmaX2-28-9B, a many-to-many multilingual MT model that performs competitively with Google Translate and GPT-4-turbo across $28$ widely spoken languages. The contributions of this paper can be summarized as follows:
\begin{itemize}
\item We benchmark the latest open-source LLMs on multilingual MT tasks in $28$ widely spoken languages, covering English-centric and Chinese-centric translation directions.
\item We systematically explore the optimal training recipe for boosting multilingual MT with LLMs and introduce the Parallel-First Monolingual-Second (PFMS) data mixing strategy for the continual pretraining stage.
\item We publicly release GemmaX2-28-9B, the many-to-many multilingual translation model supporting $28$ languages, which consistently outperforms open-source alternates and is competitive with closed-source models such as Google Translate and GPT-4-turbo.
\end{itemize}

\section{Related Work}

\subsection{Multilinguality in LLMs}

Large language models have gained substantial attention from the research community due to their outstanding performance on various tasks \cite{winata-etal-2021-language,wei2022emergentabilitieslargelanguage,touvron2023llamaopenefficientfoundation}. However, most advancements in LLMs have primarily focused on English or Chinese, resulting in inadequate performance in low-resource languages \cite{ebrahimi-etal-2022-americasnli,asai-etal-2024-buffet}. The reason is that the language distribution of the training data for LLMs is highly imbalanced and the quality varies across languages \cite{ding-etal-2024-data}. Numerous efforts have been made to enhance the multilingual capabilities of LLMs. Some studies seek to enhance the performance of LLMs on low-resource languages through continual pretraining \cite{cui2024efficienteffectivetextencoding} or supervised finetuning \cite{ustun-etal-2024-aya} using data from these languages. Additionally, some researchers \cite{li-etal-2024-improving-context} leverage contrastive learning to align the internal representations of different languages, allowing the model to improve its cross-lingual capabilities with minimal training data. In our study, we assessed the latest open-source LLMs on translation tasks across $28$ languages. We find that these latest models still show significant performance disparities across different languages, indicating that low-resource languages continue to pose a significant challenge for these models. Therefore, we expand the Gemma2 model to cover $28$ commonly used languages, enabling it to demonstrate strong generative and translation capabilities across these languages.

\subsection{Multilingual MT with LLMs}

The use of LLMs has shown significant progress in MT tasks \cite{lu2024llamaxscalinglinguistichorizons,DBLP:conf/icml/XuSCTSDM024,li-etal-2024-eliciting,gao2024boostingmanytomanymultilingualmachine}, leading to different approaches on LLM-based translation. One line of work focuses on the in-context translation capabilities, and LLMs are provided with parallel sentences to guide the model in generating the target sentence. Some studies \cite{agrawal-etal-2023-context,zhu-etal-2024-towards-robust,cui-etal-2024-efficiently} show that MT performance can be improved by using semantically related parallel sentences as examples, exhibiting promising results in scenarios with limited computational resources and insufficient parallel data.

Another line involves finetuning with translation instructions. \citet{DBLP:conf/iclr/Xu0SA24} initially pretrained the base model on monolingual data, followed by finetuning on small human-written parallel datasets, which resulted in strong translation performance. \citet{guo-etal-2024-novel} challenge the perspective that the impact of parallel data is reduced in the LLM era and demonstrate the effectiveness of parallel data during continual pretraining on enhancing multilingual translation. However, the distribution of the parallel datasets could be highly imbalanced across different languages for massively multilingual translation scenarios. \citet{alves2024tower} incorporates a fixed multilingual mixture of monolingual (two-thirds) and parallel (one-third) data, further enhancing LLMs' translation capabilities. Our work differs in that we primarily focus on exploring the optimal mixing strategy of monolingual and parallel data during continual pretraining to achieve the best translation performance.

\section{Datasets and Baseline Settings}

In this section, we describe the evaluation datasets used in Sections \ref{sec:benchmark_mt} and \ref{sec:gemmax} as well as the model configurations.

\begin{figure*}[h]
\centering
\includegraphics[scale=0.315]{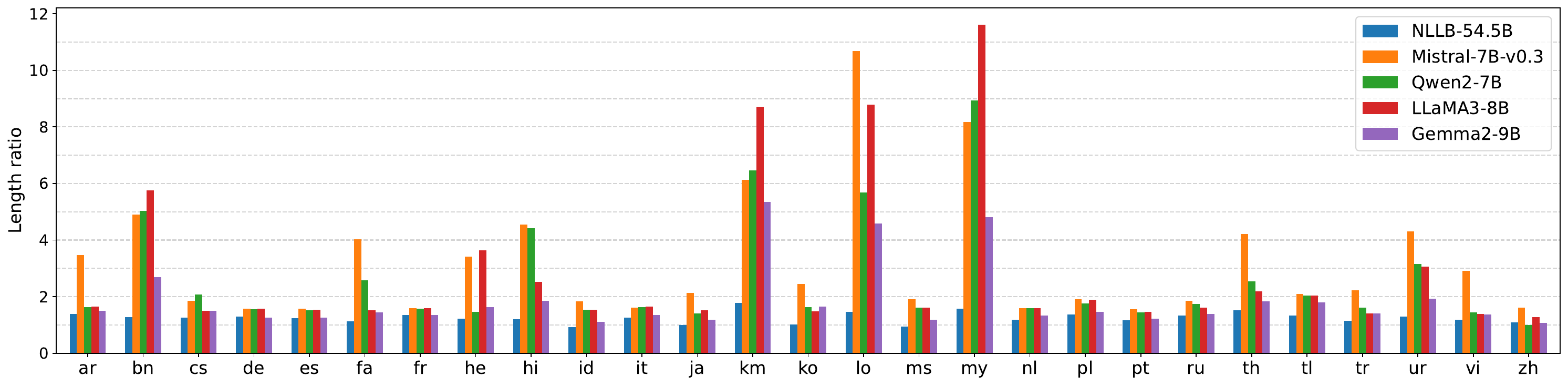} 
\caption{The tokenizer efficiency of open-source LLMs for each non-English language. The smaller the length ratio is, the more efficient the tokenizer is.}
\label{fig:tokenization}
\end{figure*}

\subsection{Datasets}

We conduct experiments on $28$ languages across a broad linguistic spectrum. The detailed information of all languages is summarized in Table \ref{tab:langs}. We evaluate the multilingual translation performance on the FLORES-200 \cite{nllbteam2022languageleftbehindscaling} benchmark. To avoid data leakage issues, we also consider the WMT-24 test sets for evaluating multilingual translation capabilities. Only sources are provided in the WMT-24 benchmark, and we adopt English sentences for reference-free evaluation.

\subsection{Models}\label{sec:baselines}

We evaluate the translation performance of six open-source LLMs: Mistral-7B-v0.3 \cite{jiang2023mistral7b}, Qwen2/2.5-7B \cite{yang2024qwen2technicalreport}, LLaMA3/3.1-8B \cite{dubey2024llama3herdmodels} and Gemma2-9B \cite{gemmateam2024gemma2improvingopen}. We also summarize the results of several SOTA models as follows:
\begin{itemize}[leftmargin=*]
\item Google Translate: The translation performance by leveraging the Google Translate API\footnote{\url{https://translate.google.com/}}. We include it to represent the commercial MT system.
\item GPT-3.5/4 Turbo: The translation performance by leveraging the OpenAI API\footnote{\url{https://api.openai.com/v1/chat/completions}} with greedy decoding for inference. Notably, we adopt the same in-context learning strategy as that employed by open-source LLMs.
\item NLLB-54.5B: The largest and best multilingual NMT model with encoder-decoder architecture released by the No Language Left Behind (NLLB) project \cite{nllbteam2022languageleftbehindscaling}.
\end{itemize}

\subsection{Evaluation}

For the FLORES-200 benchmark, we evaluate MT performance by spBLEU\footnote{We calculate the spBLEU scores via sacreBLEU \cite{post-2018-call} with the \texttt{flores200} tokenizer.} \cite{goyal-etal-2022-flores} and COMET\footnote{\url{https://huggingface.co/Unbabel/wmt22-comet-da}} \cite{rei-etal-2020-comet}. For the WMT-24 benchmark, we adopt two reference-free models, XCOMET\footnote{\url{https://huggingface.co/Unbabel/XCOMET-XXL}} \cite{guerreiro2023xcomettransparentmachinetranslation} and COMETKiwi\footnote{\url{https://huggingface.co/Unbabel/wmt23-cometkiwi-da-xxl}} \cite{rei-etal-2023-scaling}, each of which has $10$B parameters and demonstrates high correlation with human judgments \cite{freitag-etal-2023-results}.

\begin{table*}[h]\small
\centering
\begin{tabular}{l | c | c c c c} 
\hline
\multicolumn{1}{c|}{Model} & \multicolumn{1}{c|}{WMT-24} & \multicolumn{4}{c}{FLORES-200} \\
& \texttt{en} $\rightarrow$ \texttt{xx} & \texttt{en} $\rightarrow$ \texttt{xx} & \texttt{xx} $\rightarrow$ \texttt{en} & \texttt{zh} $\rightarrow$ \texttt{xx} & \texttt{xx} $\rightarrow$ \texttt{zh} \\
\hline
\hline
Google Translate & 77.64 / 73.00 & 41.52 / 88.51 & 45.35 / 88.83 & 29.64 / 85.69 & 34.86 / 87.20 \\
GPT-4-turbo & 77.55 / 70.68 & 37.01 / 87.14 & 41.53 / 88.40 & 24.66 / 84.31 & 28.80 / 85.90 \\
GPT-3.5-turbo & 70.63 / 62.64 & 32.24 / 82.49 & 36.18 / 85.63 & 20.35 / 79.34 & 24.12 / 82.61 \\
NLLB-54.5B & - & 37.34 / 87.05 & 43.63 / 88.32 & 25.41 / 84.42 & 20.73 / 80.72 \\
\hline
Gemma2-9B & 72.06 / 67.05 & 33.05 / 84.65 & 42.00 / 87.94 & 20.54 / 80.54 & 27.77 / 85.34 \\
Llama3.1-8B & 65.23 / 58.89 & 28.52 / 81.60 & 37.72 / 86.74 & 17.66 / 78.09 & 25.53 / 83.49 \\
Llama3-8B & 65.02 / 58.10 & 27.93 / 81.28 & 37.39 / 86.59 & 16.83 / 77.40 & 22.81 / 83.11 \\
Qwen2.5-7B & 59.36 / 51.59 & 22.65 / 75.20 & 34.64 / 84.94 & 14.97 / 72.64 & 25.30 / 83.16 \\
Qwen2-7B & 58.12 / 49.68 & 21.69 / 75.68 & 34.22 / 85.15 & 14.44 / 73.19 & 25.45 / 83.55 \\
Mistral-7B-v0.3 & 51.21 / 40.48 & 17.04 / 67.61 & 29.44 / 81.35 & 8.18 / 62.31 & 12.19 / 72.62 \\
\end{tabular}
\caption{Performance of different models on WMT-24 (XCOMET / COMETKiwi) and FLORES-200 (spBLEU / COMET) benchmarks. The detailed experimental results are summarized in Tables \ref{tab:wmt24_en}, \ref{tab:flores200_en}, and \ref{tab:flores200_zh}.}
\label{tab:main_results_in-context}
\end{table*}

\section{Benchmarking Open LLMs for Multilingual Machine Translation}\label{sec:benchmark_mt}

In this section, we report empirical results on tokenizer efficiency and multilingual MT performance across various open-source LLMs.

\subsection{Tokenizer Efficiency}

The open-source LLMs are typically pretrained on one or a few dominant languages, and tokenization on low-resource languages usually results in undesirable long sequences of subwords which leads to excessive GPU memory consumption during training and slow decoding speed during inference.

Following the line of \citet{liao2024ikunwmt24generalmt}, we compare the length differences between tokenized English sentences and their corresponding non-English counterparts to evaluate the tokenizer efficiency across different LLMs. Specifically, we define the length ratio as follows:
\begin{equation}
\text{length ratio} = \frac{\text{length}(\text{tokenizer}(y))}{\text{length}(\text{tokenizer}(x))},
\end{equation}
where $x$ represents the English sentence, and $y$ denotes the corresponding non-English sentence. A smaller length ratio is preferred, meaning that the tokenizer encodes the non-English sentences as efficiently as the English sentence.

We conduct our experiments on the FLORES-200 devtest dataset, which includes $1012$ sentences for each language. Note that LLaMA3 and LLaMA3.1 share the same tokenizer, as do Qwen2 and Qwen2.5. We also include the tokenizer of the NLLB-54.5B model as a strong baseline for comparison. The averaged length ratio for each language is illustrated in Figure \ref{fig:tokenization}. By checking the length ratio for various languages, we have the following observations: 1) Due to its extensive support and optimization for more than $200$ languages, NLLB-54.5B has the most balanced tokenizer and achieves low length ratios across different languages. 2) Open-source LLMs demonstrate notably high length ratios for low-resource languages such as Khmer, Lao, and Burmese. Gemma2-9B consistently achieves better length ratios for almost all languages compared with other LLMs.

\subsection{In-context Multilingual Translation Performance with Open LLMs}

\begin{figure}[h]
\centering
\includegraphics[scale=0.44]{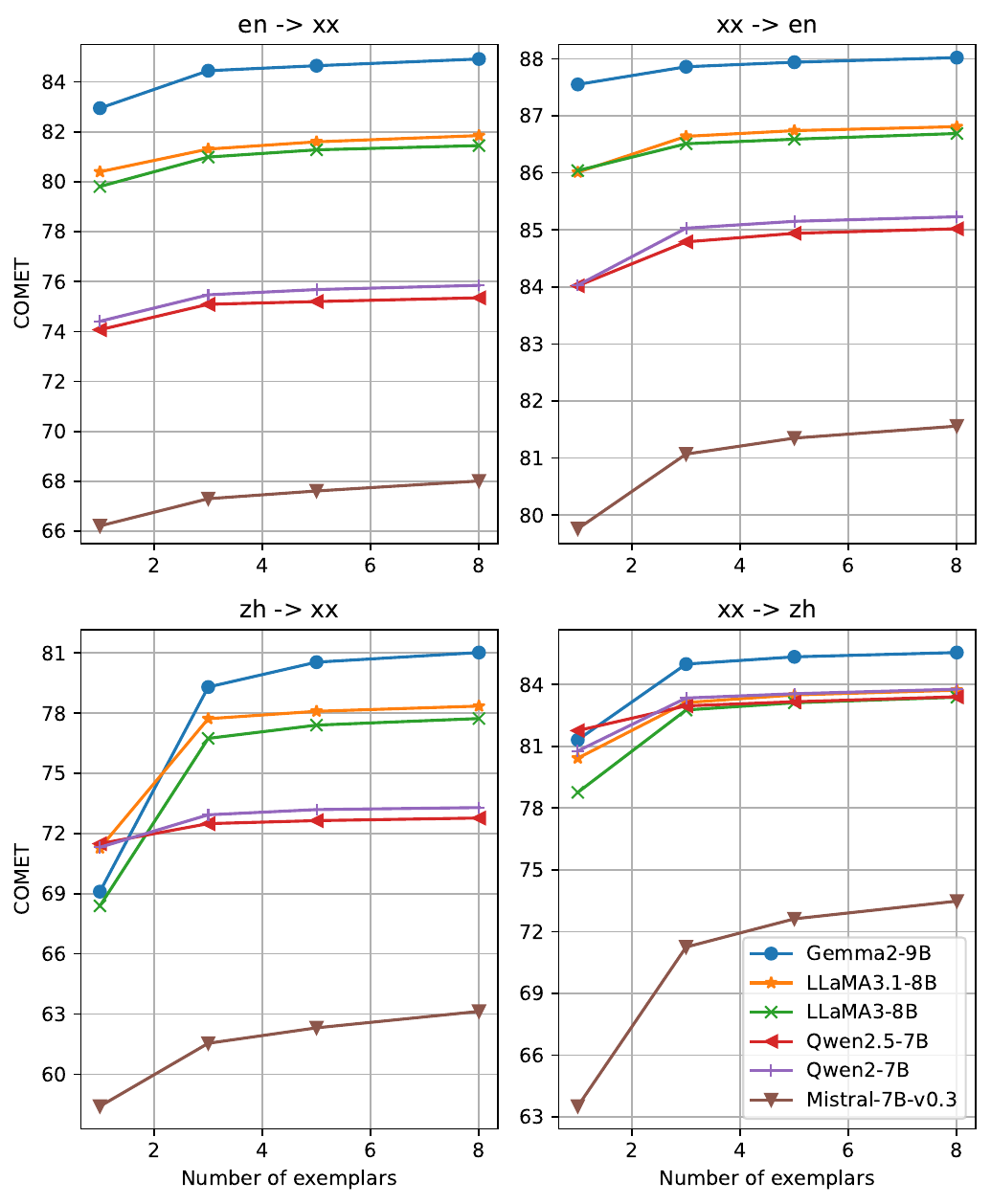} 
\caption{MT performance on the FLORES-200 benchmark with different numbers of in-context exemplars.}
\label{fig:incontext_translation}
\end{figure}

\begin{figure*}[h]
\centering
\includegraphics[scale=0.34]{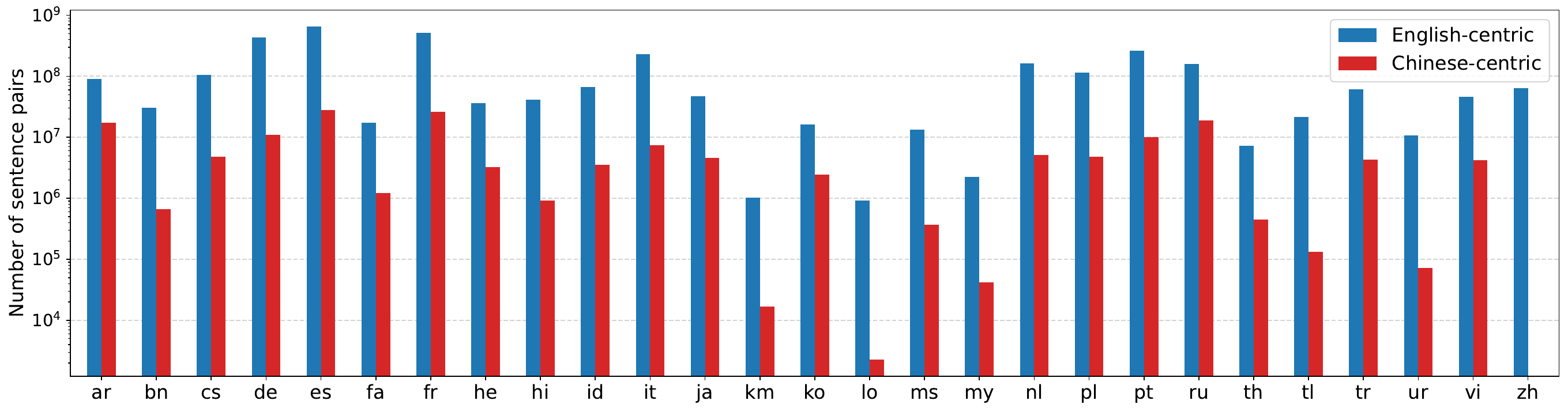} 
\caption{Number of sentences in different languages for Chinese-centric and English-centric parallel dataset.}
\label{fig:parallel_dataset}
\end{figure*}

We here investigate the multilingual translation performance of different LLMs with in-context learning strategies on the FLORES-200 and WMT-24 benchmarks. For each open LLM, we report its translation performance across $28$ languages with five randomly selected translation pairs\footnote{We apply the same five randomly selected translation pairs as exemplars for each direction during evaluation.} from the FLORES-200 development dataset as the in-context exemplars. Note that we also adopt the FLORES-200 development dataset for the WMT-24 benchmark. Following the line of \citet{zhu-etal-2024-multilingual}, we adopt the format ``<X>=<Y>'' as the in-context template, where <X> and <Y> denote the source and target sentences of the select parallel sentence pairs. All experiments are conducted based on OpenICL\footnote{\url{https://github.com/Shark-NLP/OpenICL}} \cite{wu-etal-2023-openicl}.

We report the averaged multilingual translation performance on the WMT-24 and FLORES-200 benchmarks in Tables \ref{tab:main_results_in-context}. By checking the translation performance for various languages and models, we have the following observations:
\begin{itemize}[leftmargin=*]
\item Open LLMs demonstrate impressive multilingual translation capability. Specifically, Gemma2-9B performs the best among the open LLMs evaluated and outperforms GPT-3.5-turbo on average.
\item Open LLMs still lag behind the strong supervised NMT models such as NLLB-54.5B, especially in low-resource languages.
\item Google Translate exhibits remarkable multilingual translation performance and even outperforms GPT-4-turbo on both the FLORES-200 and WMT-24 benchmarks. 
\item We do not observe serious data leakage issues on public datasets (FLORES-200) for the open LLMs we evaluated, where the translation performance on both WMT-24 and FLORES-200 benchmarks share a similar trend.
\end{itemize}

We then illustrate the translation performance on the FLORES-200 benchmark according to varying numbers of in-context exemplars in Figure \ref{fig:incontext_translation}. We can see that the COMET scores increase dramatically with the number of exemplars increasing from $1$ to $5$, and the translation performance plateaus afterward except for Mistral-7B-v0.3 which has relatively weak multilingual capability.

\section{GemmaX: Boosting Multilingual Translation with Gemma Models}\label{sec:gemmax}

Given its impressive multilingual capabilities discussed in Section \ref{sec:benchmark_mt}, we select Gemma2-9B as our backbone model for learning many-to-many multilingual machine translation across $28$ languages. We continue the pretraining of Gemma2-9B with multilingual corpora and finetune the continual pretrained model using a small but high-quality parallel dataset with the translation prompt as follows: Translate this from [\textit{source language}] to [\textit{target language}]:\textbackslash n[\textit{source language}]: <\textit{source sentence}>\textbackslash n[\textit{target language}]:<\textit{target sentence}>.

\subsection{Pretraining Data}

\paragraph{Monolingual Data} We collect monolingual data from CulturaX \cite{nguyen-etal-2024-culturax} and MADLAD-400 \cite{kudugunta2023madlad400multilingualdocumentlevellarge}. CulturaX contains $6.3$ trillion tokens in $167$ languages, which undergo meticulous cleaning and deduplication through a comprehensive pipeline. MADLAD-400 is a multilingual dataset with $3$ trillion tokens in $419$ languages, on which the authors perform a self-audit to guarantee the data quality.

\begin{figure*}[h]
\centering
\includegraphics[scale=0.45]{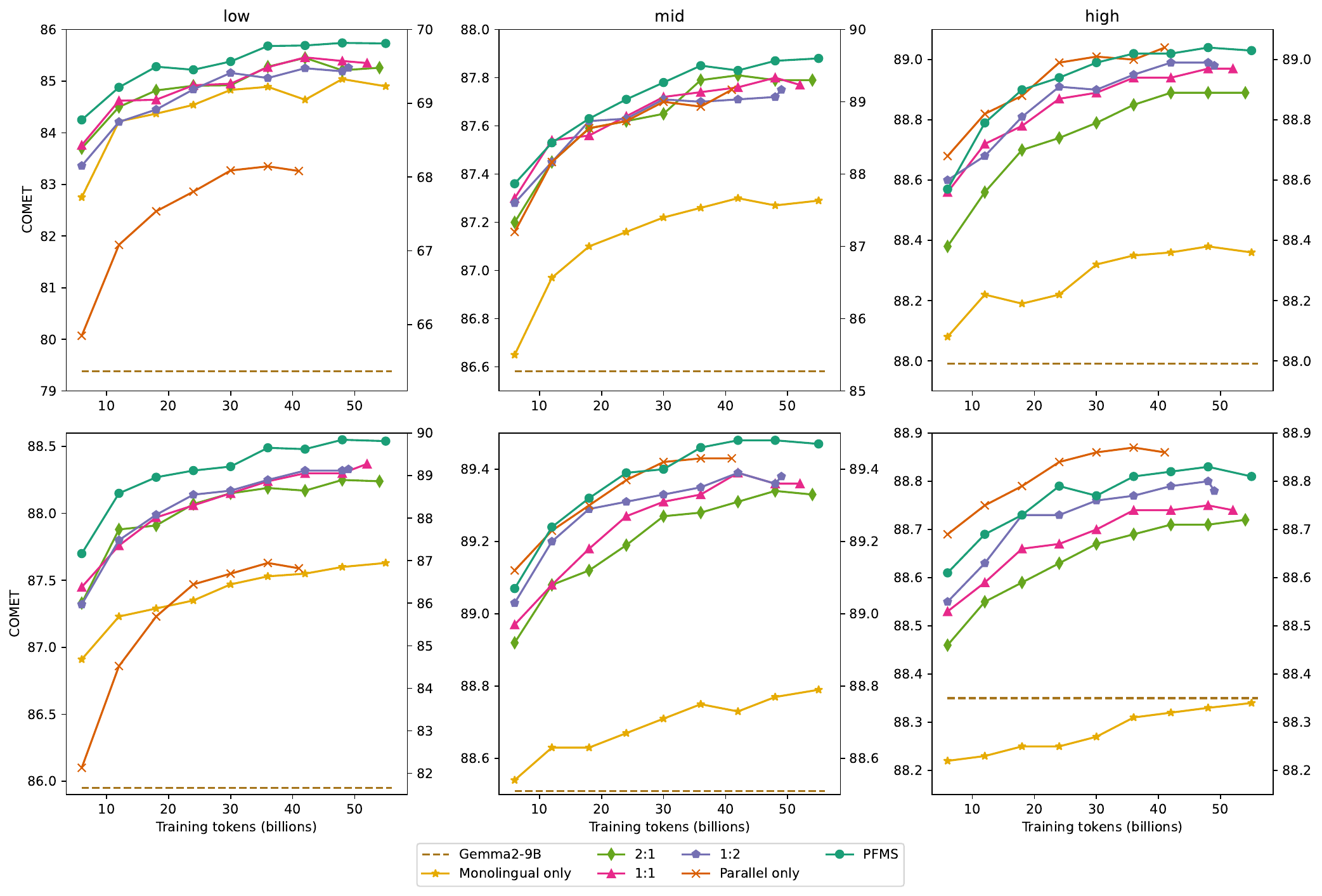} 
\caption{The translation performance (COMET) of models trained with different data recipes during continual pretraining on low-resource (left), mid-resource (middle), and high-resource (right) languages. The upper subfigures illustrate the \texttt{en}$\rightarrow$\texttt{xx} translation performance, while the lower subfigures depict the \texttt{xx}$\rightarrow$\texttt{en} translation performance. Note that ``Gemma2-9B'' refers to the direct finetuning of the model without continual pretraining, and its performance is reflected in the right-hand y-axis. The translation performance in BLEU scores is illustrated in Figure \ref{fig:ratio_bleu}.}
\label{fig:ratio}
\end{figure*}

\paragraph{Parallel Data}
We collect all Chinese-centric and English-centric parallel datasets from the OPUS collection\footnote{\url{http://www.opus.nlpl.eu}} \cite{TIEDEMANN12.463} up to August 2024, which is comprised of multiple corpora, ranging from movie subtitles \cite{tiedemann-2016-finding} to Bible \cite{christodouloupoulos2015massively} to web crawled datasets \cite{el-kishky-etal-2020-ccaligned,schwenk-etal-2021-ccmatrix}. We download all available corpora and concatenate them without curating the datasets or trying to balance the representation of different domains. After collecting all parallel datasets, we adopt the data-cleaning process as follows: 1) We remove duplicate sentence pairs and discard sentence pairs by utilizing some heuristic approaches. 2) Language identification filtering is applied by utilizing the fastText toolkit \cite{joulin2016fasttextzipcompressingtextclassification,joulin-etal-2017-bag}. 3) Semantic similarity filtering is performed based on LaBSE \cite{feng-etal-2022-language} and MuSR \cite{gao-etal-2023-learning-multilingual}. Specifically, we only keep the sentence pairs that have similarity scores between $0.75$ and $0.99$. After the cleaning process, we have about 3.4 billion cleaned Chinese-centric and English-centric sentence pairs covering 28 languages. The distribution of our parallel datasets for each language is illustrated in Figure \ref{fig:parallel_dataset}.

\subsection{Supervised Finetuning Data}

Inspired by \citet{DBLP:conf/iclr/Xu0SA24} that a small amount of high-quality data could dramatically boost the translation performance, we construct our finetuning dataset mostly from human-annotated records: 1) We extract the sentence pairs covering ten languages from the general translation task in the TowerBlock dataset\footnote{\url{https://huggingface.co/datasets/Unbabel/TowerBlocks-v0.2}}. 2) For the languages not covered in TowerBlock, we randomly sample $1000$ English-centric sentence pairs for each language either from the NTREX-128 \cite{federmann-etal-2022-ntrex} and FLORES-200 dev datasets or the OPUS dataset bidirectionally filtered by \textit{wmt23-cometkiwi-da-xxl} model with quality scores above $0.85$. 3) To mitigate the off-target issue and enhance the zero-shot translation performance, inspired by \citet{wu-etal-2024-far}, we also randomly select $25$ non-English-centric translation directions and sample $100$ sentence pairs for each direction from the NTREX-128 and FLORES-200 dev datasets. In summary, the total number of sentence pairs is around $196$K, where there are about $189$K in the English-centric directions and $7$K in other directions.

\subsection{Training Configuration}

We train all models with the LLaMA-Factory \cite{zheng-etal-2024-llamafactory} framework for one epoch on $32$ and $8$ NVIDIA H800 GPUs for the pretraining and finetuning stages respectively.

In the pretraining stage, we utilize an effective batch size of 1.57 million tokens, corresponding to a batch size of $4$ per GPU with a gradient accumulation of $6$ and a maximum sequence length of $2048$. We use the AdamW optimizer with a learning rate of 2e-5. We adopt full-weight pretraining with bf16 precision by employing a cosine learning rate scheduler. In the finetuning stage, we adopt a batch size of $4$ per GPU with a gradient accumulation of $8$ and a maximum sequence length of $2048$. Additionally, we employ an inverse square root learning rate scheduler. Detailed training configurations can be found in Tables \ref{tab:pretrain_setup} and \ref{tab:sft_setup}. We use greedy decoding to generate translations.

\subsection{Exploring the Best Data Recipe for Multilingual Translation with LLMs}\label{data_recipe}

\citet{DBLP:conf/iclr/Xu0SA24} conduct continual pretraining solely on monolingual data followed by instruction finetuning on a small number of high-quality translation pairs, achieving impressive translation results. They conclude that the reliance on parallel data is diminished in the era of LLMs. Subsequently, \citet{guo-etal-2024-novel} reemphasize the importance of parallel data and propose a three-stage training strategy: 1) continual pretraining with monolingual data; 2) continual pretraining with parallel data; 3) finetuning with source language consistent translation instructions. However, as illustrated in Figure \ref{fig:parallel_dataset}, the parallel data adopted in the second stage could be highly imbalanced for massively multilingual settings, which would weaken the model's understanding and generative capabilities on low-resource languages learned in the first stage. \citet{alves2024tower} utilize a multilingual mixture of monolingual (two-thirds) and parallel (one-third) data during continual pretraining, which significantly improves the translation performance of their model. In our experiments, we follow the idea of a multilingual mixture in the continual pretraining stage and raise two questions: 1) \textit{whether a monolingual to parallel ratio of 2:1 is the best data recipe for multilingual machine translation?} 2) \textit{If not, what is the best data recipe?}

Therefore, we consider the following configurations when preparing the pretraining dataset: 1) monolingual data only, 2) $2:1$ ratio of monolingual to parallel data, 3) $1:1$ ratio, 4) $1:2$ ratio, and 5) parallel data only. For each language, we collect two billion tokens, allocating them between monolingual and parallel sentences based on the specified ratios mentioned above. For each parallel sentence, we construct a new sentence in either <\textit{source sentence}>\textbackslash n<\textit{target sentence}> or <\textit{target sentence}>\textbackslash n<\textit{source sentence}> manner, with the order of the source and target sentences determined randomly. We randomly intersperse the monolingual and parallel data. Due to the inherent characteristics of low-resource languages, the corresponding monolingual or parallel data may be insufficient, resulting in the total number of tokens in the pretraining dataset falling short of $56$ billion. As a result, we construct five pretraining datasets corresponding to these configurations.


We continually pretrain the Gemma2-9B model using five datasets containing varying proportions of monolingual and parallel data, followed by translation instruction finetuning on our high-quality datasets. Based on Table \ref{tab:langs}, we classify the $28$ languages into high-resource ($18$), mid-resource ($7$), and low-resource ($3$) languages according to their resource availability. We then report the translation performance during continual pretraining on these three language groups in Figures \ref{fig:ratio} and \ref{fig:ratio_bleu}.

By checking model performance under different combinations of data recipes and language groups, we have the following observations:
\begin{itemize}[leftmargin=*]
\item The current prevalent training paradigm, which involves continual pretraining on large-scale monolingual data, proves to be suboptimal. While this approach improves the overall performance across various languages, it does so at the cost of translation quality for high-resource languages, particularly in the \texttt{xx}$\rightarrow$\texttt{en} directions.
\item Incorporating parallel data into the pretraining stage, regardless of the volume, significantly enhances the model's overall translation capabilities. Notably, the translation performance for high-resource language directions improves as the quantity of parallel data increases. However, such improvement was not evident for low-resource and mid-resource languages.
\item We believe that the base model has sufficient understanding and generation capabilities for high-resource languages. It mainly needs parallel data during pretraining to align the internal representations across different languages. In contrast, the model's generative and comprehension abilities for low-resource and mid-resource languages are not up to par. To improve its performance with these languages, it's crucial to include more data in the pretraining process to enhance its generative capacities. Unfortunately, due to the nature of low-resource and mid-resource languages, we are unable to obtain enough parallel data. Therefore, we think it's essential to incorporate monolingual data into the pretraining dataset for low-resource and mid-resource languages.
\end{itemize}

\begin{table*}[h!]\small
\centering
\begin{tabular}{l | c | c c c c} 
\hline
\multicolumn{1}{c|}{Model} & \multicolumn{1}{c|}{WMT-24} & \multicolumn{4}{c}{FLORES-200} \\
& \texttt{en} $\rightarrow$ \texttt{xx} & \texttt{en} $\rightarrow$ \texttt{xx} & \texttt{xx} $\rightarrow$ \texttt{en} & \texttt{zh} $\rightarrow$ \texttt{xx} & \texttt{xx} $\rightarrow$ \texttt{zh} \\
\hline
\hline
\textbf{$10$ languages} & & \\
TowerInstruct-7B-v0.2 & 84.48 / 75.02 & 38.91 / 88.46 & 42.20 / 88.29 & 23.36 / 85.21 & 26.87 / 85.69 \\
TowerInstruct-13B-v0.1 & 86.10 / 76.74 & 40.60 / 88.89 & 43.14 / 88.50 & 25.28 / 86.13 & 29.32 / 86.59 \\
GemmaX2-28-2B & 84.50 / 74.42 & 40.15 / 88.56 & 42.03 / 88.27 & 25.71 / 85.85 & 31.91 / 87.04 \\
GemmaX2-28-9B & \textbf{86.59} / \textbf{77.25} & \textbf{42.58} / \textbf{89.15} & \textbf{43.95} / \textbf{88.62} & \textbf{28.55} / \textbf{86.84} & \textbf{34.10} / \textbf{87.70} \\
\hline
\hline
\textbf{$20$ languages} & & \\
Aya-23-8B & 77.72 / 69.55 & 34.95 / 86.81 & 39.56 / 86.57 & 23.68 / 84.28 & 25.33 / 85.15 \\
Aya-Expanse-8B & 80.15 / 74.11 &  37.57 / 87.89  & 40.48 / 87.04 & 25.39 / 85.48 & 28.21 / 86.30 \\
GemmaX2-28-2B & 80.66 / 73.07 & 39.53 / 88.51 & 42.77 / 88.44 & 25.57 / 85.54 & 31.69 / 86.81 \\
GemmaX2-28-9B & \textbf{82.95} / \textbf{75.82} & \textbf{42.27} / \textbf{89.21} & \textbf{45.23} / \textbf{88.92} & \textbf{28.68} / \textbf{86.67} & \textbf{34.43} / \textbf{87.59} \\
\hline
\hline
\textbf{$23$ languages} & & \\
X-ALMA & 81.67 / 74.73 & 39.31 / 88.69 & 42.09 / 88.54 & - & - \\
GemmaX2-28-2B & 79.83 / 72.77 & 39.32 / 88.27 & 42.72 / 88.46 & 25.54 / 85.24 & 31.42 / 86.73 \\
GemmaX2-28-9B & \textbf{82.05} / \textbf{75.31} & \textbf{41.98} / \textbf{88.97} & \textbf{45.27} / \textbf{88.97} & 28.64 / 86.36 & 34.26 / 87.56 \\
\hline
\hline
\textbf{$28$ languages} & & \\
Aya-101 & 71.67 / 63.88 & 27.70 / 84.80 & 33.84 / 86.21 & 17.36 / 81.56 & 18.70 / 81.72 \\
LLaMAX3-Alpaca-8B & 67.29 / 60.21 & 28.11 / 83.12 & 35.81 / 87.05 & 18.17 / 80.78 & 21.08 / 83.38 \\
GPT-3.5-turbo & 70.63 / 62.64 & 32.24 / 82.49 & 36.18 / 85.63 & 20.35 / 79.34 & 24.12 / 82.61 \\
NLLB-54.5B & - & 37.34 / 87.05 & 43.63 / 88.32 & 25.41 / 84.42 & 20.73 / 80.72 \\
GPT-4-turbo & 77.55 / 70.68 & 37.01 / 87.14 & 41.53 / 88.40 & 24.66 / 84.31 & 28.80 / 85.90 \\
Google Translate & 77.64 / 73.00 & \textbf{41.52} / \textbf{88.51} & \textbf{45.35} / 88.83 & \textbf{29.64} / \textbf{85.69} & \textbf{34.86} / 87.20 \\
GemmaX2-28-2B & 77.20 / 71.99 & 37.00 / 87.54 & 42.16 / 88.32 & 24.27 / 84.44 & 30.59 / 86.41 \\
GemmaX2-28-9B & \textbf{79.37} / \textbf{74.41} & 39.72 / 88.35 & 45.07 / \textbf{88.95} & 27.48 / \textbf{85.69} & 33.74 / \textbf{87.38} \\
\hline
\end{tabular}
\caption{Translation performance on WMT24 (XCOMET / COMETKiwi) and FLORES-200 (spBLEU / COMET) benchmarks. The detailed results are summarized in Tables \ref{tab:wmt24_ours}, \ref{tab:flores200_en_ours} and \ref{tab:flores200_zh_ours}.}
\label{tab:main_results}
\end{table*}

\subsection{Main Result}

Based on the experimental results in Section \ref{data_recipe}, we propose a Parallel-First Monolingual-Second (PFMS) data mixing strategy, where we give higher priority to parallel data than monolingual data when preparing the continual pretraining dataset. Specifically, for 2 billion tokens per language, we utilize parallel data as much as possible and supplement it with monolingual data if needed. The detailed information on the number of training tokens across different languages adopted in the PFMS strategy is summarized in Table \ref{tab:data_size}. The translation performance by leveraging the PFMS strategy is also illustrated in Figures \ref{fig:ratio} and \ref{fig:ratio_bleu}. We can see that the PFMS strategy consistently outperforms other data mixing strategies except for high-resource languages to English directions. Such a phenomenon might be due to the fact that the dominant part of the parallel-only dataset is the English-centric corpora with high-resource languages.

We continually pretrain Gemma2-9B with PFMS strategy and learn GemmaX2-28-9B by finetuning the pretrained model with high-quality translation pairs covering $28$ languages. Furthermore, to investigate the impact of our PFMS strategy on models of different scales, we also conduct experiments on the Gemma2-2B model. The experimental results are summarized in Table \ref{tab:main_results}. Besides the models discussed in Section \ref{sec:baselines}, we include several of the strongest open-source multilingual models as baselines, including:
\begin{itemize}[leftmargin=*]
\item TowerInstruct-7/13B \cite{alves2024tower}: 7B and 13B LLaMA2-based models for translation-related tasks supporting $10$ languages.
\item  Aya-23-8B / Aya-Expanse-8B \cite{dang2024ayaexpansecombiningresearch}: 8B models with highly advanced multilingual capabilities, supporting 23 commonly used languages. Note that we only evaluate their multilingual translation performance on languages overlapping with GemmaX2.
\item X-ALMA \cite{xu2024xalmaplugplay}: a LLaMA2-based multilingual MT model covering $50$ languages, which consists of a 13B dense model and multiple language-specific modules, with a total of approximately $29$B parameters. Note that X-ALMA only supports English-centric directions, and we report the translation performance for the overlapping directions.
\item Aya-101 \cite{ustun-etal-2024-aya}: a 13B massively multilingual generative language model that follows instructions in $101$ languages.
\item LLaMAX3-Alpaca-8B \cite{lu2024llamaxscalinglinguistichorizons}: an 8B LLaMA3-based model supporting $102$ languages with powerful multilingual capabilities without loss instruction-following capabilities.
\end{itemize}

Despite having limited parameters and supporting a large number of languages, GemmaX2-28-9B consistently outperforms the current SOTA open-source models, demonstrating the effectiveness of our proposed PFMS data mixing strategy and the powerful translation capabilities of our model. In addition, our model achieves comparable translation performance to GPT-4-turbo and Google Translate, indicating that our model has achieved translation capabilities on par with industry standards. We also observe that GemmaX2-28-2B demonstrates strong multilingual translation capabilities with minimal model parameters, indicating the effectiveness of our PFMS data mixing strategy for models with varying parameter scales.

\section{Conclusion}

In this paper, we systematically evaluate the multilingual in-context translation of the latest open LLMs at a practical scale. While these models demonstrate strong translation capabilities, they still fall short compared to closed-source models. Furthermore, we explore the best data recipe for multilingual translation with LLMs and propose a Parallel-First Monolingual-Second (PFMS) data mixing strategy. Leveraging the PFMS strategy during continual pretraining on Gemma2-9B, we achieve significant improvements, bringing its translation performance to a level comparable with GPT-4-turbo and Google Translate. To the best of our knowledge, GemmaX2-28-9B is the open model with the highest translation quality. We aim to develop models that support a broader range of languages and possess enhanced translation capabilities as part of our future work.

\section*{Limitations}

Due to limited computational resources, we only conduct multilingual in-context translation evaluation and explore the optimal data mixing strategy on open LLMs with parameter sizes below ten billion. The translation performance and optimal data recipe for larger models remain unclear.




\appendix

\section{Appendix}
\label{sec:appendix}

\begin{table*}[h]
\centering
\begin{tabular}{c | c | c | c | c | c}
ISO Code & Language & Script & Family & Subgrouping & Resource \\
\hline
\hline
\texttt{ar} & Arabic & Arabic & Afro-Asiatic & Semitic & High \\
\texttt{bn} & Bengali & Bengali & Indo-European & Indo-Aryan & Mid \\
\texttt{cs} & Czech & Latin &  Indo-European & Balto-Slavic & High \\
\texttt{de} & German & Latin & Indo-European & Germanic & High \\
\texttt{en} & English &  Latin & Indo-European &  Germanic & High \\
\texttt{es} & Spanish & Latin & Indo-European & Italic & High \\
\texttt{fa} & Persian & Arabic & Indo-European & Iranian & High \\
\texttt{fr} & French &  Latin & Indo-European &  Italic & High \\
\texttt{he} & Hebrew & Hebrew & Afro-Asiatic & Semitic & Mid \\
\texttt{hi} & Hindi & Devanagari &  Indo-European & Indo-Aryan & High \\
\texttt{id} & Indonesian &  Latin &  Austronesian & Malayo-Polynesian & Mid \\
\texttt{it} & Italian & Latin & Indo-European &  Italic & High \\
\texttt{ja} & Japanese & Japanese &  Japonic &  Japanesic & High \\
\texttt{km} & Khmer & Khmer & Austroasiatic & Khmeric & Low \\
\texttt{ko} & Korean & Hangul &  Koreanic &  Korean & High \\
\texttt{lo} & Lao & Lao & Tai-Kadai & Kam-Tai & Low \\
\texttt{ms} & Malay &  Latin & Austronesian & Malayo-Polynesian  & Mid \\
\texttt{my} & Burmese & Myanmar & Sino-Tibetan & Burmo-Qiangic & Low \\
\texttt{nl} & Dutch & Latin &  Indo-European &  Germanic & High \\
\texttt{pl} & Polish &  Latin & Indo-European & Balto-Slavic & High \\
\texttt{pt} & Portuguese &  Latin & Indo-European & Italic & High \\
\texttt{ru} & Russian & Cyrillic & Indo-European &  Balto-Slavic & High \\
\texttt{th} & Thai & Thai &  Tai-Kadai &  Kam-Tai & Mid \\
\texttt{tl} & Tagalog & Latin &  Austronesian & Malayo-Polynesian & Mid \\
\texttt{tr} & Turkish & Latin & Turkic &  Common Turkic & High \\
\texttt{ur} & Urdu &  Arabic & Indo-European & Indo-Aryan & Mid \\
\texttt{vi} & Vietnamese & Latin & Austroasiatic & Vietic & High \\
\texttt{zh} & Chinese & Han & Sino-Tibetan & Sinitic & High \\
\hline
\end{tabular}
\caption{$28$ languages supported by our model. The resource of each language is determined according to the taxonomy classes by \citet{joshi-etal-2020-state}.}
\label{tab:langs}
\end{table*}


\begin{table*}[h]\tiny
\centering
\begin{tabular}{c | c c c c c c | c c c} 
Direction & Mistral-7B & Qwen2-7B & Qwen2.5-7B & Llama3-8B & Llama3.1-8B & Gemma2-9B & GPT3.5-turbo & GPT4-turbo & Google \\
\hline
\hline
\texttt{en}$\rightarrow$\texttt{ar} & 41.11 / 31.29 & 59.12 / 52.88 & 68.34 / 62.36 & 65.78 / 58.86 & 66.15 / 60.33 & 73.16 / 68.69 & 75.91 / 69.33 & 79.86 / 73.98 & 77.23 / 72.53 \\
\hline
\texttt{en}$\rightarrow$\texttt{bn} & 28.37 / 18.93 & 39.30 / 36.79 & 38.51 / 35.22 & 59.02 / 57.20 & 56.73 / 54.98 & 63.46 / 62.29 & 54.49 / 44.36 & 70.77 / 56.86 & 67.36 / 66.82 \\
\hline
\texttt{en}$\rightarrow$\texttt{cs} & 59.94 / 47.11 & 53.64 / 40.42 & 57.05 / 45.94 & 67.00 / 55.98 & 67.24 / 56.96 & 76.29 / 68.19 & 77.56 / 69.19 & 81.47 / 74.05 & 77.69 / 69.08 \\
\hline
\texttt{en}$\rightarrow$\texttt{de} & 85.70 / 62.46 & 84.15 / 60.77 & 86.52 / 65.85 & 87.59 / 68.15 & 87.94 / 68.69 & 90.17 / 73.60 & 91.09 / 75.95 & 92.65 / 77.84 & 93.63 / 79.26 \\
\hline
\texttt{en}$\rightarrow$\texttt{es} & 79.67 / 66.42 & 81.22 / 68.50 & 82.19 / 69.88 & 83.19 / 70.74 & 83.07 / 70.57 & 85.25 / 74.24 & 86.66 / 75.96 & 86.79 / 76.47 & 86.32 / 75.20 \\
\hline
\texttt{en}$\rightarrow$\texttt{fa} & 28.82 / 18.50 & 47.61 / 42.86 & 44.90 / 39.93 & 65.23 / 61.14 & 65.46 / 61.87 & 75.12 / 72.24 & 68.63 / 63.90 & 76.76 / 72.20 & 73.06 / 69.65 \\
\hline
\texttt{en}$\rightarrow$\texttt{fr} & 74.59 / 63.03 & 76.15 / 65.68 & 77.95 / 67.55 & 77.14 / 66.72 & 76.47 / 66.50 & 81.13 / 72.24 & 83.48 / 74.83 & 83.94 / 76.49 & 81.47 / 73.22 \\
\hline
\texttt{en}$\rightarrow$\texttt{he} & 29.06 / 14.47 & 48.81 / 38.16 & 48.24 / 37.52 & 58.44 / 50.15 & 60.01 / 52.30 & 70.53 / 63.78 & 63.21 / 52.77 & 75.65 / 66.95 & 74.62 / 68.29 \\
\hline
\texttt{en}$\rightarrow$\texttt{hi} & 32.57 / 26.45 & 37.75 / 34.44 & 37.76 / 35.11 & 57.23 / 55.32 & 58.60 / 56.92 & 62.72 / 62.67 & 58.70 / 47.81 & 68.42 / 54.56 & 77.30 / 71.63 \\
\hline
\texttt{en}$\rightarrow$\texttt{id} & 64.89 / 49.71 & 74.99 / 61.66 & 75.48 / 65.49 & 76.96 / 66.28 & 77.13 / 66.55 & 81.35 / 71.79 & 84.71 / 75.28 & 84.43 / 76.72 & 83.11 / 76.00 \\
\hline
\texttt{en}$\rightarrow$\texttt{it} & 76.55 / 63.93 & 77.36 / 64.60 & 79.43 / 67.36 & 79.90 / 68.23 & 80.62 / 69.44 & 84.41 / 74.70 & 85.61 / 75.22 & 87.06 / 78.18 & 83.82 / 74.54 \\
\hline
\texttt{en}$\rightarrow$\texttt{ja} & 59.27 / 61.91 & 71.76 / 74.69 & 73.05 / 76.46 & 72.84 / 74.77 & 72.16 / 74.95 & 80.98 / 81.40 & 81.23 / 78.82 & 83.98 / 83.47 & 86.64 / 86.18 \\
\hline
\texttt{en}$\rightarrow$\texttt{km} & 10.94 / 2.85 & 18.58 / 20.69 & 17.36 / 18.69 & 25.18 / 29.73 & 27.12 / 32.56 & 39.10 / 49.53 & 26.80 / 23.54 & 56.63 / 55.89 & 65.88 / 72.85 \\
\hline
\texttt{en}$\rightarrow$\texttt{ko} & 59.43 / 57.74 & 67.71 / 65.51 & 71.72 / 70.04 & 73.74 / 69.83 & 72.80 / 70.55 & 76.24 / 71.73 & 79.42 / 74.55 & 84.84 / 80.85 & 80.49 / 78.07 \\
\hline
\texttt{en}$\rightarrow$\texttt{lo} & 13.64 / 2.65 & 14.30 / 5.72 & 17.22 / 7.76 & 16.58 / 9.72 & 15.85 / 9.23 & 33.34 / 32.53 & 23.91 / 15.22 & 43.33 / 37.63 & 61.67 / 63.59 \\
\hline
\texttt{en}$\rightarrow$\texttt{ms} & 60.97 / 46.86 & 70.17 / 57.46 & 70.32 / 60.10 & 72.30 / 62.10 & 72.17 / 62.45 & 76.49 / 66.87 & 82.42 / 72.85 & 82.04 / 74.49 & 76.25 / 67.71 \\
\hline
\texttt{en}$\rightarrow$\texttt{my} & 11.72 / 1.03 & 14.43 / 10.11 & 11.48 / 6.31 & 23.52 / 25.33 & 24.49 / 28.24 & 30.28 / 37.82 & 22.64 / 14.59 & 47.87 / 45.98 & 59.83 / 70.76 \\
\hline
\texttt{en}$\rightarrow$\texttt{nl} & 78.58 / 62.17 & 78.46 / 63.04 & 79.07 / 64.96 & 83.81 / 71.15 & 84.39 / 71.31 & 87.27 / 74.96 & 89.65 / 78.53 & 90.16 / 80.58 & 88.51 / 78.86  \\
\hline
\texttt{en}$\rightarrow$\texttt{pl} & 66.30 / 52.06 & 63.52 / 48.74 & 65.01 / 53.03 & 72.70 / 60.18 & 73.06 / 60.44 & 81.96 / 71.46 & 81.32 / 71.33 & 85.56 / 76.99 & 82.59 / 74.22 \\
\hline
\texttt{en}$\rightarrow$\texttt{pt} & 78.42 / 65.55 & 80.69 / 68.02 & 82.57 / 70.61 & 81.07 / 68.80 & 80.81 / 68.49 & 84.82 / 74.23 & 87.31 / 76.98 & 87.89 / 78.31 & 86.42 / 75.78 \\
\hline
\texttt{en}$\rightarrow$\texttt{ru} & 71.72 / 61.82 & 75.28 / 67.42 & 76.32 / 68.88 & 75.63 / 67.59 & 76.34 / 67.87 & 80.52 / 74.01 & 80.25 / 72.37 & 83.74 / 76.80 & 81.05 / 74.40 \\
\hline
\texttt{en}$\rightarrow$\texttt{th} & 34.48 / 25.08 & 66.46 / 63.25 & 71.25 / 67.34 & 67.95 / 62.92 & 64.96 / 59.28 & 76.44 / 72.16 & 63.23 / 53.22 & 76.65 / 68.99 & 75.48 / 71.67 \\
\hline
\texttt{en}$\rightarrow$\texttt{tl} & 46.74 / 36.71 & 43.39 / 29.25 & 38.95 / 25.96 & 57.58 / 50.65 & 57.01 / 51.54 & 67.46 / 63.14 & 72.90 / 69.51 & 75.30 / 74.09 & 71.46 / 64.89 \\
\hline
\texttt{en}$\rightarrow$\texttt{tr} & 40.88 / 31.49 & 47.60 / 42.15 & 51.90 / 49.52 & 60.78 / 57.13 & 63.12 / 60.30 & 72.00 / 69.17 & 75.14 / 70.32 & 77.41 / 75.42 & 77.22 / 75.74 \\
\hline
\texttt{en}$\rightarrow$\texttt{ur} & 24.00 / 17.08 & 25.92 / 21.18 & 25.30 / 18.12 & 44.41 / 43.99 & 49.00 / 51.48 & 57.71 / 60.42 & 54.85 / 51.51 & 68.77 / 62.92 & 65.71 / 67.00 \\
\hline
\texttt{en}$\rightarrow$\texttt{vi} & 54.44 / 42.18 & 72.71 / 64.65 & 76.98 / 70.68 & 75.09 / 67.74 & 74.17 / 68.09 & 78.98 / 73.84 & 76.79 / 70.57 & 81.01 / 76.03 & 79.35 / 75.31 \\
\hline
\texttt{en}$\rightarrow$\texttt{zh} & 69.74 / 63.57 & 78.18 / 72.73 & 77.92 / 72.35 & 74.85 / 68.23 & 74.32 / 68.26 & 78.34 / 72.52 & 79.09 / 72.65 & 80.98 / 75.69 & 82.04 / 77.64 \\
\end{tabular}
\caption{Evaluation results (XCOMET / COMETKiwi) on the WMT-24 benchmark. Note that the translation performance of the open LLMs is based on the $5$-shot in-context learning strategy.}\label{tab:wmt24_en}
\end{table*}

\begin{table*}[h]\tiny
\centering
\begin{tabular}{c | c c c c c c | c c c c} 
Direction & Mistral-7B & Qwen2-7B & Qwen2.5-7B & Llama3-8B & Llama3.1-8B & Gemma2-9B & NLLB-54B & GPT3.5-turbo & GPT4-turbo & Google \\
\hline
\hline
\texttt{ar}$\rightarrow$\texttt{en} & 29.99 / 82.55 & 38.92 / 86.46 & 40.60 / 86.94 & 41.42 / 87.01 & 41.94 / 87.01 & 46.04 / 87.94 & 48.26 / 88.13 & 41.83 / 87.77 & 46.39 / 88.44 & 50.59 / 88.73 \\
\texttt{ar}$\leftarrow$\texttt{en} & 8.66 / 63.97 & 21.76 / 78.71 & 27.11 / 83.13 & 27.65 / 82.86 & 28.63 / 83.4 & 35.87 / 86.02 & 43.03 / 87.22 & 37.23 / 86.81 & 40.79 / 87.81 & 47.80 / 88.72 \\
\hline
\texttt{bn}$\rightarrow$\texttt{en} & 17.28 / 78.07 & 28.01 / 85.6 & 27.96 / 85.75 & 32.74 / 87.25 & 32.42 / 87.15 & 38.52 / 88.61 & 42.22 / 89.31 & 29.17 / 86.37 & 39.19 / 89.32 & 43.52 / 89.62  \\
\texttt{bn}$\leftarrow$\texttt{en} & 3.06 / 44.88 & 9.1 / 66.48 & 8.14 / 64.31 & 21.85 / 80.71 & 21.58 / 80.24 & 23.24 / 82.58 & 36.04 / 87.08 & 19.96 / 78.24 & 29.39 / 86.18 & 37.49 / 86.60 \\
\hline
\texttt{cs}$\rightarrow$\texttt{en} & 41.61 / 87.91 & 40.78 / 87.52 & 42.18 / 87.87 & 43.39 / 88.23 & 43.87 / 88.39 & 46.53 / 88.96 & 45.31 / 88.67 & 45.09 / 89.02 & 46.44 / 89.28 & 48.68 / 89.34 \\
\texttt{cs}$\leftarrow$\texttt{en} & 28.17 / 85.32 & 24.17 / 81.43 & 24.87 / 82.34 & 32.41 / 87.57 & 33.35 / 87.97 & 36.75 / 90.45 & 42.38 / 91.54 & 39.65 / 91.25 & 42.33 / 92.11 & 45.95 / 91.51 \\
\hline
\texttt{de}$\rightarrow$\texttt{en} & 46.23 / 88.99 & 45.96 / 88.95 & 45.09 / 88.92 & 47.29 / 88.99 & 47.72 / 89.18 & 50.72 / 89.62 & 49.80 / 89.36 & 48.71 / 89.80 & 50.08 / 89.89 & 52.80 / 90.19 \\
\texttt{de}$\leftarrow$\texttt{en} & 34.35 / 84.6 & 33.12 / 83.78 & 34.51 / 84.81 & 38.54 / 86.06 & 39.39 / 86.2 & 45.01 / 87.88 & 46.62 / 88.10 & 47.10 / 88.52 & 48.67 / 89.09 & 48.27 / 89.29 \\
\hline
\texttt{es}$\rightarrow$\texttt{en} & 34.94 / 86.96 & 35.24 / 87.06 & 36.05 / 87.15 & 36.95 / 87.17 & 37.34 / 87.24 & 39.02 / 87.47 & 38.80 / 87.49 & 36.27 / 87.70 & 38.23 / 87.99 & 37.35 / 87.53 \\
\texttt{es}$\leftarrow$\texttt{en} & 28.8 / 84.81 & 29.26 / 85.24 & 30.14 / 85.56 & 30.55 / 85.92 & 30.92 / 85.96 & 33.45 / 86.67 & 33.05 / 86.39 & 34.25 / 87.25 & 34.41 / 87.36 & 35.23 / 87.24 \\
\hline
\texttt{fa}$\rightarrow$\texttt{en} & 24.54 / 81.42 & 33.47 / 86.21 & 33.80 / 86.12 & 37.92 / 87.55 & 38.19 / 87.58 & 43.07 / 88.65 & 44.19 / 88.70 & 37.35 / 87.83 & 41.58 / 88.82 & 44.68 / 88.90 \\
\texttt{fa}$\leftarrow$\texttt{en} & 4.49 / 48.09 & 13.49 / 72.72 & 13.41 / 70.40 & 25.68 / 83.71 & 26.18 / 84.11 & 32.19 / 87.31 & 36.13 / 87.75 & 29.30 / 85.60 & 34.85 / 88.10 & 39.72 / 88.34 \\
\hline
\texttt{fr}$\rightarrow$\texttt{en} & 47.38 / 89.01 & 47.41 / 89.02 & 47.79 / 89.15 & 49.17 / 89.01 & 49.57 / 89.19 & 51.93 / 89.63 & 51.54 / 89.44 & 49.93 / 89.70 & 51.41 / 89.88 & 53.44 / 89.89 \\
\texttt{fr}$\leftarrow$\texttt{en} & 45.66 / 85.95 & 46.32 / 86.59 & 47.23 / 86.80 & 49.43 / 87.13 & 50.12 / 87.2 & 55.12 / 88.36 & 56.16 / 88.15 & 56.53 / 89.09 & 56.59 / 89.13 & 58.98 / 89.23 \\
\hline
\texttt{he}$\rightarrow$\texttt{en} & 24.07 / 76.46 & 40.28 / 86.57 & 41.28 / 86.70 & 44.98 / 87.81 & 45.02 / 87.84 & 49.33 / 88.80 & 49.02 / 88.62 & 42.60 / 87.57 & 47.73 / 88.99 & 51.80 / 89.19 \\
\texttt{he}$\leftarrow$\texttt{en} & 3.99 / 46.57 & 17.12 / 74.23 & 15.95 / 71.86 & 30.32 / 83.23 & 30.67 / 83.55 & 34.09 / 85.84 & 46.82 / 88.71 & 33.29 / 83.78 & 41.89 / 88.58 & 48.80 / 88.88 \\
\hline
\texttt{hi}$\rightarrow$\texttt{en} & 23.34 / 82.32 & 33.57 / 87.3 & 33.87 / 87.12 & 39.15 / 88.97 & 39.25 / 88.97 & 45.56 / 89.79 & 47.26 / 90.27 & 37.57 / 88.69 & 44.75 / 90.07 & 51.52 / 90.99 \\
\texttt{hi}$\leftarrow$\texttt{en} & 6.32 / 49.27 & 11.79 / 61.07 & 12.18 / 61.34 & 25.77 / 75 & 27.56 / 75.92 & 35.33 / 79.07 & 40.55 / 81.17 & 27.71 / 76.49 & 36.37 / 81.01 & 38.94 / 83.13 \\
\hline
\texttt{id}$\rightarrow$\texttt{en} & 41.26 / 87.9 & 44.26 / 88.94 & 45.78 / 89.16 & 46.41 / 89.16 & 46.72 / 89.29 & 50.87 / 90.00 & 49.88 / 89.37 & 47.66 / 89.83 & 49.72 / 90.22 & 53.25 / 90.29 \\
\texttt{id}$\leftarrow$\texttt{en} & 24.9 / 82.61 & 34.75 / 88.5 & 37.40 / 89.19 & 41.02 / 90.06 & 41.4 / 90.15 & 48.53 / 91.55 & 49.18 / 91.18 & 47.13 / 91.80 & 49.41 / 92.29 & 52.76 / 92.61 \\
\hline
\texttt{it}$\rightarrow$\texttt{en} & 38.18 / 87.76 & 38.32 / 87.82 & 38.89 / 87.90 & 39.96 / 87.93 & 40.1 / 88.1 & 41.9 / 88.52 & 41.56 / 88.13 & 39.35 / 88.45 & 40.58 / 88.67 & 41.31 / 88.54 \\
\texttt{it}$\leftarrow$\texttt{en} & 30.19 / 86.13 & 28.35 / 85.71 & 29.08 / 86.33 & 32.98 / 87.39 & 33.63 / 87.53 & 36.39 / 88.56 & 38.27 / 88.58 & 37.80 / 89.13 & 38.77 / 89.34 & 39.86 / 89.38 \\
\hline
\texttt{ja}$\rightarrow$\texttt{en} & 26.03 / 85.79 & 30.11 / 87.43 & 31.01 / 87.64 & 30.89 / 87.59 & 30.81 / 87.5 & 34.51 / 88.24 & 34.54 / 87.92 & 30.53 / 88.24 & 33.61 / 88.68 & 38.14 / 89.36 \\
\texttt{ja}$\leftarrow$\texttt{en} & 13.25 / 82.79 & 21.12 / 88.66 & 24.30 / 89.34 & 24.45 / 89.04 & 24.54 / 89.01 & 30.4 / 91.03 & 20.07 / 89.08 & 30.50 / 90.91 & 32.46 / 91.67 & 37.08 / 92.75 \\
\hline
\texttt{km}$\rightarrow$\texttt{en} & 7.82 / 60.54 & 16.21 / 77.29 & 14.79 / 74.12 & 23.86 / 82.01 & 23.98 / 82.56 & 30.3 / 84.77 & 38.64 / 87.02 & 13.09 / 73.18 & 32.98 / 87.01 & 35.10 / 85.82 \\
\texttt{km}$\leftarrow$\texttt{en} & 0.19 / 30.09 & 1.9 / 45.18 & 1.59 / 41.26 & 4.82 / 56.06 & 5.44 / 57.08 & 7.15 / 66.08 & 22.97 / 79.86 & 4.12 / 49.19 & 17.93 / 79.92 & 27.36 / 83.37 \\
\hline
\texttt{ko}$\rightarrow$\texttt{en} & 26.42 / 85.78 & 30.3 / 87.18 & 31.78 / 87.59 & 31.27 / 87.24 & 31.75 / 87.34 & 35.96 / 88.47 & 35.42 / 88.03 & 32.12 / 88.12 & 36.07 / 89.06 & 39.31 / 89.32 \\
\texttt{ko}$\leftarrow$\texttt{en} & 10.1 / 79.37 & 15.49 / 85.14 & 17.71 / 86.23 & 19.31 / 86.46 & 20.21 / 86.98 & 23.24 / 88.1 & 26.72 / 89.47 & 24.33 / 88.83 & 28.56 / 90.29 & 31.31 / 90.24 \\
\hline
\texttt{lo}$\rightarrow$\texttt{en} & 4.92 / 53.28 & 9.64 / 64.26 & 11.03 / 64.81 & 14.57 / 71.85 & 15.17 / 72.33 & 27.13 / 80.54 & 42.40 / 87.67 & 8.29 / 64.60 & 23.41 / 79.64 & 44.05 / 88.53 \\
\texttt{lo}$\leftarrow$\texttt{en} & 0.1 / 31.6 & 0.86 / 35.42 & 0.92 / 33.22 & 1.49 / 36.92 & 1.47 / 37.9 & 5.9 / 55.2 & 29.59 / 84.10 & 2.76 / 42.76 & 11.24 / 63.45 & 29.59 / 83.19 \\
\hline
\texttt{ms}$\rightarrow$\texttt{en} & 40.32 / 87.03 & 42.12 / 87.84 & 42.88 / 87.87 & 46.05 / 88.71 & 46.37 / 88.66 & 51.42 / 89.57 & 51.20 / 89.21 & 47.75 / 89.24 & 50.22 / 89.87 & 53.29 / 89.76 \\
\texttt{ms}$\leftarrow$\texttt{en} & 20.16 / 79.33 & 25.95 / 84.98 & 25.08 / 85.03 & 35.99 / 87.29 & 36.34 / 87.29 & 41.95 / 89.15 & 45.54 / 89.01 & 37.36 / 89.39 & 43.86 / 90.14 & 47.41 / 89.91 \\
\hline
\texttt{my}$\rightarrow$\texttt{en} & 2.22 / 54.38 & 6.12 / 69.18 & 4.63 / 64.70 & 17.69 / 81.08 & 18.44 / 81.62 & 24.75 / 84.00 & 34.73 / 87.50 & 2.27 / 58.68 & 25.87 / 85.72 & 34.97 / 87.23 \\
\texttt{my}$\leftarrow$\texttt{en} & 0.09 / 31.99 & 1.39 / 42.81 & 0.74 / 36.87 & 6.63 / 64 & 7.13 / 65.03 & 6.05 / 69.63 & 17.74 / 84.58 & 1.39 / 48.20 & 12.85 / 79.26 & 24.45 / 87.50 \\
\hline
\texttt{nl}$\rightarrow$\texttt{en} & 35.02 / 86.93 & 35.43 / 87.06 & 35.24 / 87.15 & 35.87 / 86.67 & 36.32 / 87.1 & 38.14 / 87.65 & 38.94 / 87.62 & 37.82 / 88.05 & 38.86 / 88.20 & 39.68 / 87.97 \\
\texttt{nl}$\leftarrow$\texttt{en} & 27.82 / 84.88 & 26.23 / 84.2 & 26.11 / 84.45 & 30.49 / 86.64 & 31 / 86.63 & 33.48 / 87.67 & 35.61 / 87.74 & 36.79 / 88.79 & 37.48 / 88.90 & 37.58 / 88.72  \\
\hline
\texttt{pl}$\rightarrow$\texttt{en} & 33.71 / 85.8 & 33.58 / 85.71 & 33.86 / 85.75 & 34.75 / 85.85 & 34.82 / 85.9 & 37.1 / 86.76 & 36.68 / 86.35 & 35.45 / 86.66 & 36.38 / 86.97 & 38.05 / 86.93 \\
\texttt{pl}$\leftarrow$\texttt{en} & 23.58 / 84.69 & 19.87 / 81.34 & 20.61 / 82.11 & 26.18 / 86.96 & 26.14 / 87.1 & 29.78 / 89.07 & 32.54 / 89.38 & 31.75 / 89.67 & 33.43 / 90.36 & 36.69 / 90.28 \\
\hline
\texttt{pt}$\rightarrow$\texttt{en} & 50.5 / 89.19 & 51.66 / 89.29 & 52.42 / 89.43 & 52.86 / 89.34 & 53.35 / 89.46 & 56.49 / 90.06 & 55.19 / 89.57 & 54.20 / 90.04 & 55.47 / 90.15 & 57.44 / 90.30 \\
\texttt{pt}$\leftarrow$\texttt{en} & 45.43 / 87.36 & 45.59 / 88.17 & 47.61 / 88.57 & 49.34 / 88.67 & 49.46 / 88.71 & 54.04 / 89.78 & 52.87 / 89.26 & 55.85 / 90.42 & 56.95 / 90.57 & 56.98 / 90.58 \\
\hline
\texttt{ru}$\rightarrow$\texttt{en} & 38.65 / 86.25 & 38.25 / 86.33 & 39.07 / 86.63 & 39.68 / 86.37 & 39.9 / 86.48 & 42.51 / 87.06 & 42.21 / 86.98 & 39.18 / 87.04 & 41.04 / 87.27 & 43.95 / 87.52 \\
\texttt{ru}$\leftarrow$\texttt{en} & 31.17 / 86.29 & 31.25 / 86.93 & 32.42 / 87.61 & 34.36 / 87.82 & 34.88 / 87.77 & 38.61 / 89.46 & 41.03 / 89.45 & 37.86 / 89.51 & 40.71 / 90.36 & 44.35 / 90.60 \\
\hline
\texttt{th}$\rightarrow$\texttt{en} & 20.03 / 80.62 & 31.91 / 87.25 & 33.40 / 87.76 & 33.91 / 87.68 & 34.36 / 87.71 & 37.92 / 88.59 & 36.87 / 87.88 & 27.97 / 86.70 & 35.76 / 88.90 & 36.96 / 88.07 \\
\texttt{th}$\leftarrow$\texttt{en} & 6.01 / 55.55 & 27.29 / 83.73 & 31.36 / 85.11 & 31.13 / 84.93 & 30.69 / 83.82 & 37.71 / 87.72 & 35.08 / 85.71 & 31.57 / 84.16 & 40.68 / 88.65 & 46.31 / 88.77 \\
\hline
\texttt{tl}$\rightarrow$\texttt{en} & 36.99 / 82.95 & 40.03 / 84.37 & 38.55 / 83.14 & 45.97 / 86.24 & 46.36 / 86.45 & 52.44 / 88.01 & 54.58 / 88.38 & 48.54 / 87.93 & 53.96 / 88.87 & 57.18 / 88.78 \\
\texttt{tl}$\leftarrow$\texttt{en} & 11.21 / 66.19 & 11.08 / 67.32 & 8.10 / 61.47 & 25.45 / 79.77 & 26.19 / 79.67 & 33.09 / 83.87 & 38.34 / 84.79 & 33.79 / 84.99 & 39.84 / 86.13 & 39.76 / 84.79 \\
\hline
\texttt{tr}$\rightarrow$\texttt{en} & 28.22 / 84.38 & 33.75 / 86.91 & 34.72 / 87.38 & 38.42 / 88.64 & 38.73 / 88.6 & 43.68 / 89.68 & 45.82 / 89.84 & 42.29 / 89.86 & 44.80 / 90.23 & 48.79 / 90.48 \\
\texttt{tr}$\leftarrow$\texttt{en} & 8.35 / 65.91 & 13.06 / 77.01 & 18.33 / 80.25 & 24.86 / 84.88 & 25.52 / 85.26 & 34.7 / 88.63 & 41.52 / 89.84 & 37.20 / 89.53 & 40.32 / 90.58 & 45.38 / 91.25 \\
\hline
\texttt{ur}$\rightarrow$\texttt{en} & 14.07 / 73.73 & 25.92 / 83.11 & 24.44 / 82.00 & 31.89 / 85.57 & 32.82 / 85.98 & 38.73 / 87.45 & 43.08 / 88.30 & 30.82 / 85.71 & 39.96 / 88.54 & 43.26 / 88.39 \\
\texttt{ur}$\leftarrow$\texttt{en} & 1.83 / 39.69 & 3.59 / 53.02 & 2.30 / 46.69 & 13.05 / 71.02 & 16.64 / 74.04 & 20.58 / 78.4 & 30.52 / 81.75 & 20.20 / 76.07 & 28.71 / 82.92 & 32.46 / 83.05 \\
\hline
\texttt{vi}$\rightarrow$\texttt{en} & 31.87 / 84.55 & 38.52 / 87.11 & 39.39 / 87.43 & 39.79 / 87.14 & 40.49 / 87.37 & 43.55 / 88.04 & 43.75 / 87.79 & 38.48 / 87.62 & 41.87 / 88.34 & 45.40 / 88.40 \\
\texttt{vi}$\leftarrow$\texttt{en} & 18.83 / 74.06 & 34.62 / 86.56 & 37.17 / 87.84 & 38.24 / 87.46 & 38.37 / 87.66 & 42.31 / 89.01 & 43.30 / 88.19 & 39.36 / 88.50 & 43.07 / 89.63 & 47.08 / 89.90 \\
\hline
\texttt{zh}$\rightarrow$\texttt{en} & 29.33 / 85.88 & 34.2 / 87.22 & 34.80 / 87.23 & 32.66 / 86.77 & 32.66 / 86.86 & 35.96 / 87.52 & 36.13 / 87.16 & 32.60 / 87.48 & 34.97 / 87.72 & 39.82 / 88.46 \\
\texttt{zh}$\leftarrow$\texttt{en} & 23.28 / 83.37 & 37.11 / 88.43 & 37.06 / 88.36 & 32.2 / 86.99 & 32.5 / 87.04 & 37.28 / 88.42 & 26.61 / 82.25 & 35.72 / 88.42 & 37.84 / 88.96 & 43.36 / 89.94 \\
\end{tabular}
\caption{English-centric evaluation results (spBLEU / COMET) on the FLORES-200 benchmark. Note that the translation performance of the open LLMs is based on the $5$-shot in-context learning strategy.}\label{tab:flores200_en}
\end{table*}

\begin{table*}[h]\tiny
\centering
\begin{tabular}{c | c c c c c c | c c c c} 
Direction & Mistral-7B & Qwen2-7B & Qwen2.5-7B & Llama3-8B & Llama3.1-8B & Gemma2-9B & NLLB-54B & GPT3.5-turbo & GPT4-turbo & Google \\
\hline
\hline
\texttt{ar}$\rightarrow$\texttt{zh} & 6.78 / 67.75 & 26.64 / 84.24 & 27.49 / 84.64 & 22.01 / 82.54 & 22.84 / 82.66 & 28.54 / 84.87 & 21.25 / 80.25 & 25.05 / 83.20 & 29.73 / 85.48 & 36.21 / 86.50   \\
\texttt{ar}$\leftarrow$\texttt{zh} & 2.91 / 56.15 & 14.7 / 77.21 & 16.53 / 80.10 & 14.09 / 78.08 & 15.96 / 79.07 & 22.66 / 82.16 & 27.53 / 84.08 & 22.66 / 83.65 & 26.24 / 84.62 & 32.32 / 85.30 \\
\hline
\texttt{bn}$\rightarrow$\texttt{zh} & 5.16 / 65.41 & 21.86 / 82.77 & 21.24 / 82.69 & 20.77 / 82.3 & 21.35 / 82.68 & 25.39 / 84.94 & 20.01 / 81.33 & 19.40 / 81.58 & 27.38 / 86.37 & 33.55 / 87.07 \\
\texttt{bn}$\leftarrow$\texttt{zh} & 0.74 / 37.07 & 5.54 / 61.44 & 4.85 / 58.94 & 12.81 / 74.3 & 12.71 / 73.98 & 13.27 / 76.97 & 23.67 / 81.96 & 12.81 / 72.07 & 20.64 / 82.36 & 27.64 / 82.38 \\
\hline
\texttt{cs}$\rightarrow$\texttt{zh} & 17.4 / 80.41 & 28.93 / 85.82 & 28.40 / 85.64 & 25.2 / 84.51 & 25.62 / 84.74 & 30.42 / 86.25 & 19.60 / 79.15 & 29.11 / 86.06 & 31.30 / 86.62 & 36.19 / 87.51 \\
\texttt{cs}$\leftarrow$\texttt{zh} & 11.15 / 75.53 & 14.45 / 79.42 & 14.07 / 80.08 & 18.13 / 85.07 & 19.24 / 86.15 & 22.17 / 87.39 & 25.63 / 88.81 & 24.22 / 89.02 & 26.90 / 90.42 & 30.65 / 89.81 \\
\hline
\texttt{de}$\rightarrow$\texttt{zh} & 19.32 / 81.31 & 30.96 / 86.8 & 30.68 / 86.81 & 27.33 / 85.5 & 27.42 / 85.51 & 31.82 / 86.93 & 21.48 / 80.87 & 30.38 / 86.82 & 31.65 / 87.07 & 37.03 / 87.97 \\
\texttt{de}$\leftarrow$\texttt{zh} & 16.15 / 77.83 & 20.08 / 80.72 & 21.06 / 81.75 & 21.2 / 82.14 & 22.08 / 82.5 & 24.65 / 83.03 & 27.15 / 84.27 & 27.06 / 84.84 & 29.73 / 85.93 & 32.97 / 86.27 \\
\hline
\texttt{en}$\rightarrow$\texttt{zh} & 23.28 / 83.37 & 37.11 / 88.43 & 37.06 / 88.36 & 32.2 / 86.99 & 32.5 / 87.04 & 37.28 / 88.42 & 26.61 / 82.25 & 35.72 / 88.42 & 37.97 / 88.51 & 43.36 / 89.94 \\
\texttt{en}$\leftarrow$\texttt{zh} & 29.33 / 85.88 & 34.2 / 87.22 & 34.80 / 87.23 & 32.66 / 86.77 & 32.66 / 86.86 & 35.96 / 87.52 & 36.13 / 87.16 & 32.60 / 87.48 & 35.23 / 87.78 & 39.82 / 88.46 \\
\hline
\texttt{es}$\rightarrow$\texttt{zh} & 16.91 / 81.54 & 27.93 / 86.63 & 28.40 / 86.76 & 24.26 / 85.33 & 24.77 / 85.61 & 28.02 / 86.81 & 19.03 / 79.85 & 27.60 / 86.63 & 28.42 / 86.74 & 32.74 / 87.24 \\
\texttt{es}$\leftarrow$\texttt{zh} & 15.06 / 80.17 & 20.56 / 83.72 & 20.93 / 83.94 & 20.09 / 83.44 & 20.65 / 83.64 & 23.26 / 84.64 & 23.69 / 84.46 & 23.40 / 85.35 & 24.04 / 85.35 & 25.59 / 85.47 \\
\hline
\texttt{fa}$\rightarrow$\texttt{zh} & 9.14 / 71.62 & 24.98 / 84.44 & 24.60 / 84.16 & 23.02 / 83.81 & 23.15 / 84.16 & 27.38 / 85.66 & 21.89 / 81.66 & 24.32 / 84.47 & 28.84 / 86.17 & 34.43 / 87.16 \\
\texttt{fa}$\leftarrow$\texttt{zh} & 1.67 / 43.92 & 9.43 / 71 & 8.21 / 68.32 & 14.84 / 79.92 & 16.61 / 81.01 & 21.16 / 83.97 & 23.10 / 84.43 & 18.45 / 82.34 & 23.34 / 85.83 & 27.68 / 85.42 \\
\hline
\texttt{fr}$\rightarrow$\texttt{zh} & 19.01 / 81.3 & 31.89 / 87 & 30.97 / 86.91 & 26.6 / 85.4 & 27.07 / 85.6 & 31.56 / 86.88 & 21.19 / 80.44 & 30.31 / 86.79 & 32.32 / 87.44 & 37.71 / 87.90 \\
\texttt{fr}$\leftarrow$\texttt{zh} & 22.32 / 79.96 & 28.14 / 83.33 & 28.62 / 83.24 & 26.7 / 82.68 & 27.68 / 83.04 & 28.42 / 81.98 & 34.16 / 84.28 & 32.94 / 85.22 & 34.58 / 85.73 & 38.24 / 85.89 \\
\hline
\texttt{he}$\rightarrow$\texttt{zh} & 7.65 / 66.82 & 26.08 / 83.58 & 25.99 / 83.70 & 23.36 / 82.85 & 24.38 / 83.12 & 29.31 / 85.17 & 19.77 / 78.50 & 24.92 / 83.29 & 30.47 / 85.66 & 36.89 / 86.80 \\
\texttt{he}$\leftarrow$\texttt{zh} & 1.09 / 41.27 & 8.92 / 71.81 & 7.65 / 69.60 & 16.25 / 78.76 & 16.4 / 79.35 & 18.61 / 81.91 & 27.04 / 84.87 & 17.26 / 79.40 & 24.25 / 85.04 & 31.91 / 85.77 \\
\hline
\texttt{hi}$\rightarrow$\texttt{zh} & 7.18 / 69.31 & 24.34 / 83.71 & 23.66 / 83.41 & 23.35 / 83.75 & 23.58 / 84.09 & 27.86 / 85.47 & 21.12 / 81.60 & 23.27 / 83.75 & 29.00 / 86.48 & 35.64 / 87.51 \\
\texttt{hi}$\leftarrow$\texttt{zh} & 1.93 / 40.23 & 6.12 / 52.79 & 7.02 / 54.20 & 14.35 / 67.57 & 14.97 / 67.84 & 17.84 / 70.81 & 24.20 / 73.81 & 16.06 / 69.22 & 22.60 / 74.71 & 28.80 / 75.99 \\
\hline
\texttt{id}$\rightarrow$\texttt{zh} & 16.48 / 79.47 & 30.22 / 86.13 & 30.57 / 86.24 & 26.25 / 84.81 & 26.79 / 85.12 & 31.15 / 86.49 & 21.38 / 79.92 & 28.83 / 86.10 & 31.36 / 86.56 & 37.24 / 87.66 \\
\texttt{id}$\leftarrow$\texttt{zh} & 7.72 / 75.08 & 20.73 / 85.59 & 22.22 / 86.64 & 21.6 / 86.24 & 22.62 / 86.77 & 26.44 / 87.2 & 29.55 / 87.99 & 25.70 / 88.21 & 28.86 / 89.22 & 33.18 / 89.40 \\
\hline
\texttt{it}$\rightarrow$\texttt{zh} & 17.95 / 81.48 & 29.17 / 87.01 & 29.20 / 86.70 & 24.87 / 85.33 & 25.51 / 85.47 & 29.14 / 86.83 & 19.33 / 79.68 & 28.32 / 86.74 & 29.77 / 87.17 & 34.59 / 87.65 \\
\texttt{it}$\leftarrow$\texttt{zh} & 14.88 / 80.28 & 19.26 / 84.04 & 19.42 / 84.48 & 20.1 / 84.58 & 21.04 / 84.9 & 23.9 / 85.94 & 25.77 / 85.73 & 25.47 / 86.99 & 27.39 / 87.63 & 28.98 / 87.31 \\
\hline
\texttt{ja}$\rightarrow$\texttt{zh} & 15.36 / 81.36 & 25.74 / 87.51 & 26.62 / 87.54 & 21.79 / 85.61 & 22.32 / 85.9 & 26.06 / 87.4 & 19.64 / 81.81 & 24.94 / 87.42 & 26.09 / 87.16 & 33.54 / 88.71 \\
\texttt{ja}$\leftarrow$\texttt{zh} & 8.58 / 81.35 & 16.65 / 87.72 & 18.59 / 88.68 & 17.42 / 87.32 & 18.15 / 87.69 & 21.51 / 89.08 & 16.78 / 87.83 & 22.44 / 89.74 & 24.42 / 90.76 & 28.21 / 90.97 \\
\hline
\texttt{km}$\rightarrow$\texttt{zh} & 1.26 / 49.34 & 13.74 / 76.05 & 10.52 / 71.66 & 14.78 / 78.29 & 16.12 / 78.85 & 20.82 / 82.03 & 21.33 / 83.00 & 9.16 / 69.56 & 23.85 / 84.61 & 29.39 / 85.41 \\
\texttt{km}$\leftarrow$\texttt{zh} & 0.14 / 28.15 & 1.47 / 42.66 & 1.38 / 39.96 & 3.25 / 51.52 & 3.81 / 52.31 & 4.51 / 60.19 & 17.55 / 79.83 & 2.83 / 46.69 & 14.22 / 77.03 & 22.26 / 80.68 \\
\hline
\texttt{ko}$\rightarrow$\texttt{zh} & 14.79 / 79.61 & 26.7 / 86.41 & 27.12 / 86.50 & 23.4 / 84.71 & 24.23 / 85.17 & 26.74 / 86.19 & 21.31 / 83.01 & 25.00 / 85.82 & 28.56 / 86.89 & 34.03 / 87.87 \\
\texttt{ko}$\leftarrow$\texttt{zh} & 4.39 / 73.84 & 11.77 / 83.31 & 13.79 / 85.00 & 14.25 / 84.34 & 15.65 / 85.14 & 16.98 / 85.8 & 20.39 / 86.86 & 17.77 / 86.67 & 21.65 / 88.40 & 24.80 / 88.12 \\
\hline
\texttt{lo}$\rightarrow$\texttt{zh} & 0.94 / 46.14 & 6.6 / 62.99 & 8.62 / 63.95 & 8.02 / 68.08 & 9.25 / 69.03 & 16.84 / 77.22 & 23.44 / 83.71 & 5.57 / 61.33 & 16.28 / 76.24 & 33.45 / 86.67 \\
\texttt{lo}$\leftarrow$\texttt{zh} & 0.09 / 30.12 & 0.51 / 34.27 & 0.58 / 32.13 & 0.74 / 34.22 & 0.89 / 34.62 & 2.67 / 47.97 & 24.30 / 82.08 & 1.50 / 41.01 & 7.66 / 59.23 & 22.51 / 80.22 \\
\hline
\texttt{ms}$\rightarrow$\texttt{zh} & 15.32 / 78.42 & 27.99 / 84.8 & 28.33 / 84.74 & 24.98 / 83.76 & 26.02 / 84.16 & 30.16 / 85.66 & 23.16 / 80.89 & 28.11 / 85.11 & 30.49 / 85.89 & 36.75 / 87.02 \\
\texttt{ms}$\leftarrow$\texttt{zh} & 6.28 / 72 & 14.14 / 82.04 & 14.67 / 82.71 & 18.48 / 83.34 & 19.16 / 83.82 & 21.86 / 84.93 & 26.33 / 85.77 & 18.77 / 85.32 & 23.98 / 86.76 & 30.23 / 86.50 \\
\hline
\texttt{my}$\rightarrow$\texttt{zh} & 0.21 / 43.3 & 5.8 / 69.34 & 3.99 / 63.66 & 12.51 / 77.49 & 13.76 / 78.56 & 16.98 / 80.35 & 16.77 / 80.42 & 1.27 / 55.77 & 18.91 / 82.18 & 27.72 / 85.02 \\
\texttt{my}$\leftarrow$\texttt{zh} & 0.02 / 29.98 & 1.2 / 40.85 & 0.68 / 34.90 & 4.72 / 56.48 & 5.16 / 58.1 & 3.88 / 62.3 & 15.79 / 82.04 & 1.20 / 45.56 & 10.39 / 75.62 & 20.33 / 84.40 \\
\hline
\texttt{nl}$\rightarrow$\texttt{zh} & 15.9 / 80.08 & 27.38 / 85.72 & 27.21 / 85.61 & 23.79 / 84.21 & 24.22 / 84.54 & 27.29 / 85.8 & 16.61 / 77.64 & 26.87 / 85.91 & 28.46 / 86.32 & 32.10 / 86.66 \\
\texttt{nl}$\leftarrow$\texttt{zh} & 13.09 / 77.7 & 17.57 / 81.87 & 17.77 / 82.17 & 19.2 / 82.99 & 19.78 / 83.7 & 21.48 / 84.03 & 24.74 / 85.31 & 23.69 / 85.53 & 25.89 / 86.64 & 28.00 / 86.50 \\
\hline
\texttt{pl}$\rightarrow$\texttt{zh} & 16.04 / 79.7 & 26.73 / 85.3 & 26.58 / 85.14 & 22.87 / 83.61 & 23.81 / 84.04 & 27.42 / 85.44 & 17.37 / 77.75 & 25.96 / 85.12 & 28.00 / 85.76 & 32.27 / 86.52 \\
\texttt{pl}$\leftarrow$\texttt{zh} & 13.15 / 78.46 & 15.14 / 81.76 & 14.70 / 81.59 & 17.6 / 85.21 & 18.1 / 85.73 & 21 / 86.86 & 23.19 / 87.82 & 22.94 / 88.56 & 25.06 / 89.37 & 27.95 / 89.29 \\
\hline
\texttt{pt}$\rightarrow$\texttt{zh} & 18.81 / 81.45 & 31.29 / 87.27 & 31.02 / 87.15 & 26.51 / 85.63 & 27.47 / 85.94 & 31.31 / 87.25 & 20.54 / 80.02 & 30.22 / 86.57 & 32.78 / 87.46 & 38.01 / 88.43 \\
\texttt{pt}$\leftarrow$\texttt{zh} & 19.15 / 81.99 & 25.5 / 85.6 & 27.19 / 85.78 & 25.14 / 85.13 & 25.98 / 85.54 & 29.62 / 86.11 & 31.17 / 86.39 & 29.93 / 87.28 & 32.44 / 87.66 & 34.84 / 87.67 \\
\hline
\texttt{ru}$\rightarrow$\texttt{zh} & 15.22 / 78.65 & 29.07 / 85.74 & 29.43 / 85.88 & 24.57 / 83.81 & 25.1 / 84.1 & 29.61 / 85.43 & 21.22 / 80.67 & 27.55 / 85.34 & 29.77 / 85.75 & 35.13 / 86.64 \\
\texttt{ru}$\leftarrow$\texttt{zh} & 13.36 / 75.88 & 20.64 / 85.83 & 22.11 / 86.42 & 20.91 / 85.79 & 21.69 / 86.52 & 24.22 / 86.68 & 28.21 / 87.99 & 24.58 / 88.01 & 27.23 / 88.47 & 31.52 / 88.99 \\
\hline
\texttt{th}$\rightarrow$\texttt{zh} & 4.48 / 65.73 & 26.43 / 86.56 & 27.00 / 86.69 & 22.56 / 84.69 & 23.2 / 85.15 & 27.49 / 86.46 & 20.07 / 82.01 & 20.47 / 83.73 & 28.23 / 87.27 & 31.78 / 87.03 \\
\texttt{th}$\leftarrow$\texttt{zh} & 2.15 / 55.72 & 21.29 / 81.02 & 24.81 / 83.01 & 24.55 / 82.23 & 22.19 / 80.98 & 29.39 / 85 & 30.20 / 84.27 & 24.51 / 80.97 & 33.01 / 86.53 & 38.26 / 86.75 \\
\hline
\texttt{tl}$\rightarrow$\texttt{zh} & 13.26 / 73.48 & 25.35 / 81.22 & 23.90 / 80.34 & 22.66 / 80.99 & 23.82 / 81.47 & 29.7 / 84.14 & 21.61 / 79.12 & 25.58 / 82.62 & 31.03 / 84.90 & 37.18 / 86.02 \\
\texttt{tl}$\leftarrow$\texttt{zh} & 3.51 / 60.23 & 4.86 / 62.35 & 3.51 / 57.10 & 10.59 / 73.37 & 12.24 / 74.61 & 19.13 / 80.13 & 22.81 / 80.87 & 19.59 / 81.25 & 23.29 / 82.36 & 25.30 / 81.40 \\
\hline
\texttt{tr}$\rightarrow$\texttt{zh} & 13.63 / 76.11 & 26.32 / 83.84 & 26.51 / 84.42 & 23.87 / 83.59 & 24.54 / 83.95 & 28.63 / 85.39 & 20.27 / 79.47 & 27.21 / 85.15 & 30.30 / 86.38 & 36.06 / 87.42 \\
\texttt{tr}$\leftarrow$\texttt{zh} & 3.58 / 59.25 & 10.44 / 74.55 & 10.88 / 75.85 & 13.37 / 79.38 & 16.72 / 80.6 & 20.64 / 82.8 & 24.82 / 84.47 & 21.99 / 84.55 & 23.67 / 85.79 & 30.19 / 86.85 \\
\hline
\texttt{ur}$\rightarrow$\texttt{zh} & 2.77 / 58.04 & 19.76 / 80 & 18.69 / 79.35 & 20.07 / 81.06 & 21.24 / 81.99 & 24.59 / 83.9 & 21.79 / 82.21 & 20.06 / 81.23 & 27.07 / 85.45 & 33.02 / 86.07  \\
\texttt{ur}$\leftarrow$\texttt{zh} & 0.59 / 34.17 & 2.14 / 48.52 & 1.17 / 40.56 & 6.72 / 63.72 & 8.66 / 68.09 & 11.31 / 73.18 & 20.39 / 78.26 & 11.72 / 70.27 & 19.34 / 79.05 & 23.44 / 78.96 \\
\hline
\texttt{vi}$\rightarrow$\texttt{zh} & 14.77 / 79.62 & 28.11 / 86.63 & 29.17 / 86.72 & 24.38 / 85.22 & 25.15 / 85.62 & 28.37 / 86.69 & 21.79 / 82.34 & 26.03 / 86.01 & 29.70 / 86.75 & 35.14 / 87.95 \\
\texttt{vi}$\leftarrow$\texttt{zh} & 7.88 / 70.03 & 24.52 / 85.44 & 26.99 / 87.00 & 24.78 / 85.28 & 25.94 / 85.99 & 28.15 / 86.12 & 31.75 / 87.63 & 27.31 / 87.10 & 29.76 / 88.12 & 34.69 / 88.74 \\
\end{tabular}
\caption{Chinese-centric evaluation results (spBLEU / COMET) on the FLORES-200 benchmark. Note that the translation performance of the open LLMs is based on the $5$-shot in-context learning strategy.}\label{tab:flores200_zh}
\end{table*}

\begin{table}[h]\small
\centering
\begin{tabular}{l|l}
\hline
Training batch size on each device & 4 \\
Number of GPUs & 32 \\
Gradient Accumulation Steps & 6 \\
Maximum Sequence Length & 2048 \\
Number of Epochs & 1 \\
Learning Rate & 2e-5\\
LR Scheduler & cosine \\
Finetuning Type & full \\
bf16 & true \\
Template & empty \\
Maximum Gradient Norm & 1.0 \\
Warmup & 0.01 \\
Weight Decay & 0.01 \\
Optimizer & AdamW \\
DeepSpeed & ZeRO2 \\
\hline
\end{tabular}
\caption{Hyperparameter settings for the pretraining experiments.}
\label{tab:pretrain_setup}
\end{table}

\begin{table}[h]\small
\centering
\begin{tabular}{l|l}
\hline
Training batch size on each device & 4 \\
Number of GPUs & 8 \\
Gradient Accumulation Steps & 8 \\
Maximum Sequence Length & 2048 \\
Number of Epochs & 1 \\
Learning Rate & 2e-5\\
LR Scheduler & inverse sqrt \\
Finetuning Type & full \\
bf16 & true \\
Template & empty \\
Maximum Gradient Norm & 1.0 \\
Warmup & 0.01 \\
Weight Decay & 0.01 \\
Optimizer & AdamW \\
DeepSpeed & ZeRO2 \\
\hline
\end{tabular}
\caption{Hyperparameter settings for the finetuing experiments.}
\label{tab:sft_setup}
\end{table}

\begin{figure*}[h]
\centering
\includegraphics[scale=0.45]{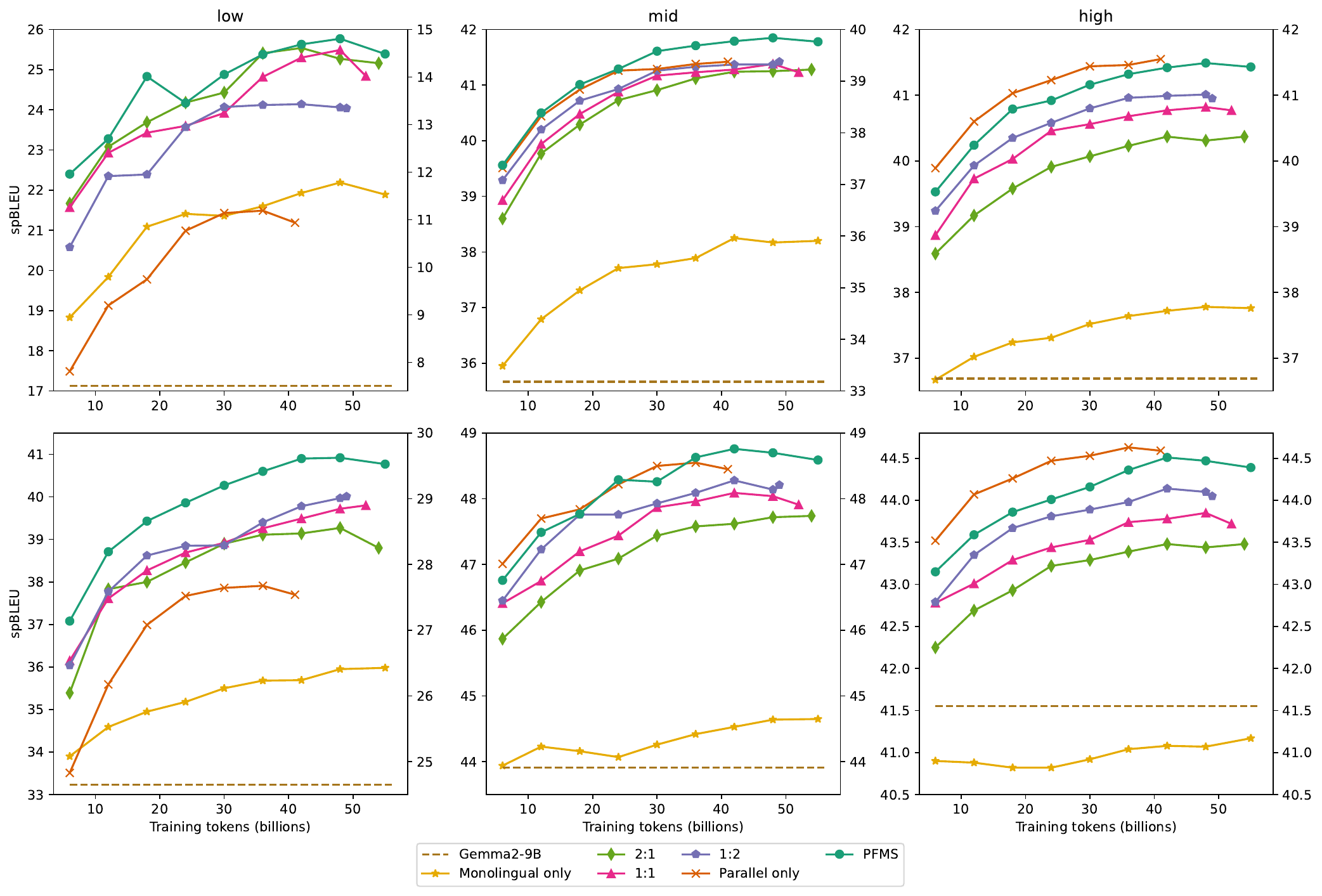} 
\caption{The translation performance (BLEU) of models trained with different data recipes during continual pretraining on low-resource (left), mid-resource (middle), and high-resource (right) languages. The upper subfigures illustrate the \texttt{en}$\rightarrow$\texttt{xx} translation performance, while the lower subfigures depict the \texttt{xx}$\rightarrow$\texttt{en} translation performance. Note that ``Gemma2-9B'' refers to the direct finetuning of the model without continual pretraining, and its performance is reflected in the right-hand y-axis.}
\label{fig:ratio_bleu}
\end{figure*}

\begin{table*}[h]\small
    \centering
    \begin{tabular}{c|c|c|c}
        Languages & \# Monolingual tokens & \# Parallel tokens (English-centric) & \# Parallel tokens (Chinese-centric) \\
        \hline
        \hline
        \texttt{ar} & 0 & 1000000000 & 1000000000 \\ \hline
        \texttt{bn} & 504825391 & 1452758417 & 42416192 \\ \hline
        \texttt{cs} & 0 & 1712873789 & 287126211 \\ \hline
        \texttt{de} & 0 & 1325095947 & 674904053 \\ \hline
        \texttt{en} & 0 & - & 2000000000 \\ \hline
        \texttt{es} & 0 & 1000000000 & 1000000000 \\ \hline
        \texttt{fa} & 1114273750 & 823160488 & 62565762 \\ \hline
        \texttt{fr} & 0 & 1000000000 & 1000000000 \\ \hline
        \texttt{he} & 332751438 & 1526487076 & 140761486 \\ \hline
        \texttt{hi} & 0 & 1932427027 & 67572973 \\ \hline
        \texttt{id} & 0 & 1800539057 & 199460943 \\ \hline
        \texttt{it} & 0 & 1506228911 & 493771089 \\ \hline
        \texttt{ja} & 0 & 1679229006 & 320770994 \\ \hline
        \texttt{km} & 1883737183 & 113850449 & 2412368 \\ \hline
        \texttt{ko} & 815056806 & 964279235 & 220663959 \\ \hline
        \texttt{lo} & 1154721722 & 68003946 & 262157 \\ \hline
        \texttt{ms} & 1435182152 & 544982172 & 19835676 \\ \hline
        \texttt{my} & 1832116262 & 160466198 & 7417540 \\ \hline
        \texttt{nl} & 0 & 1676682297 & 323317703 \\ \hline
        \texttt{pl} & 0 & 1720056384 & 279943616 \\ \hline
        \texttt{pt} & 0 & 1385756006 & 614243994 \\ \hline
        \texttt{ru} & 0 & 1000000000 & 1000000000 \\ \hline
        \texttt{th} & 1461701620 & 511027417 & 27270963 \\ \hline
        \texttt{tl} & 1090609718 & 898064981 & 11325301 \\ \hline
        \texttt{tr} & 0 & 1765499728 & 234500272 \\ \hline
        \texttt{ur} & 1393220838 & 600203213 & 6575949 \\ \hline
        \texttt{vi} & 0 & 1714651057 & 285348943 \\ \hline
        \texttt{zh} & 0 & 2000000000 & - \\ \hline
    \end{tabular}
    \caption{Statistics of all datasets used in the PFMS strategy.}
    \label{tab:data_size}
\end{table*}

\begin{table*}[h]\tiny
    \centering
    \resizebox{1\textwidth}{!}{
        \begin{tabular}{c|ccccccc}
        Direction & TowerInstruct-7B & TowerInstruct-13B & X-ALMA        & Aya-101       & LLaMAX3-Alpaca-8B & GemmaX2-28-2B & GemmaX2-28-9B    \\ \hline \hline
        \texttt{en}$\rightarrow$\texttt{ar}     & -                     & -                      & 78.72 / 73.56 & 67.77 / 62.85 & 65.41 / 58.95  & 75.26 / 70.21 & 78.16 / 73.06 \\ \hline
        \texttt{en}$\rightarrow$\texttt{bn}     & -                     & -                      & -             & 64.62 / 60.78 & 58.3 / 56.21   & 70.93 / 70.36 & 70.53 / 70.86 \\ \hline
        \texttt{en}$\rightarrow$\texttt{cs}     & -                     & -                      & 80.1 / 72.76  & 69.62 / 61.35 & 66.97 / 57.09  & 77.75 / 69.11 & 81.48 / 73.48 \\ \hline
        \texttt{en}$\rightarrow$\texttt{de}     & 91.05 / 74.12         & 91.99 / 76.28          & 92.64 / 76.6  & 86.69 / 65.64 & 86.34 / 65.96  & 90.84 / 73.67 & 92.42 / 76.63 \\ \hline
        \texttt{en}$\rightarrow$\texttt{es}     & 86.46 / 75.03         & 87.44 / 75.97          & 86.93 / 76.16 & 80.22 / 67.42 & 81.36 / 67.9   & 86.45 / 74.22 & 88.05 / 76.96 \\ \hline
        \texttt{en}$\rightarrow$\texttt{fa}     & -                     & -                      & 78.59 / 74.49 & 68.46 / 65.64 & 65.79 / 62.31  & 76.22 / 72.41 & 79.84 / 75.87 \\ \hline
        \texttt{en}$\rightarrow$\texttt{fr}     & 83.01 / 73.74         & 84.02 / 75.33          & 83.05 / 75.1  & 74.04 / 62.09 & 74.28 / 63.05  & 82.07 / 72.28 & 84.46 / 75.66 \\ \hline
        \texttt{en}$\rightarrow$\texttt{he}     & -                     & -                      & 79.68 / 72.82 & 66.34 / 58.6  & 63.9 / 56.43   & 77.87 / 71.03 & 81.12 / 74.65 \\ \hline
        \texttt{en}$\rightarrow$\texttt{hi}     & -                     & -                      & 70.12 / 64.37 & 61.31 / 54.32 & 56.35 / 55.55  & 65.08 / 64.38 & 66.18 / 66.03 \\ \hline
        \texttt{en}$\rightarrow$\texttt{id}     & -                     & -                      & 84.53 / 76.62 & 77.79 / 67.8  & 76.17 / 63.08  & 82.22 / 73.63 & 83.58 / 74.95 \\ \hline
        \texttt{en}$\rightarrow$\texttt{it}     & 85.87 / 75.57         & 86.91 / 76.71          & 85.53 / 76.12 & 77.5 / 63.45  & 77.78 / 66.23  & 85.69 / 74.16 & 87.1 / 76.89  \\ \hline
        \texttt{en}$\rightarrow$\texttt{ja}     & -                     & -                      & 82.54 / 81.23 & 73.32 / 73.39 & 71.6 / 74.13   & 79.59 / 79.01 & 81.81 / 80.82 \\ \hline
        \texttt{en}$\rightarrow$\texttt{km}     & -                     & -                      & -             & 67.42 / 72.55 & 45.71 / 52.32  & 65.64 / 71.71 & 68.21 / 74.38 \\ \hline
        \texttt{en}$\rightarrow$\texttt{ko}     & 80.87 / 77.51         & 83.83 / 80.01          & 82.03 / 77.65 & 68.96 / 65.74 & 74.22 / 69.81  & 80.52 / 77.34 & 84.73 / 80.8  \\ \hline
        \texttt{en}$\rightarrow$\texttt{lo}     & -                     & -                      & -             & 69.68 / 67.24 & 43.31 / 32.9   & 62.3 / 64.41 & 64.35 / 66.5  \\ \hline
        \texttt{en}$\rightarrow$\texttt{ms}     & -                     & -                      & 81.13 / 73.29 & 76.69 / 61.22 & 72.73 / 61.18  & 78.96 / 69.98 & 78.29 / 69.64 \\ \hline
        \texttt{en}$\rightarrow$\texttt{my}     & -                     & -                      & -             & 58.72 / 67.7  & 37.44 / 41.92  & 61.72 / 71.46 & 67.37 / 75.74 \\ \hline
        \texttt{en}$\rightarrow$\texttt{nl}     & 89.81 / 78.88         & 90.75 / 79.7           & 90.2 / 79.38  & 83.16 / 68.3  & 81.09 / 68.13  & 88.89 / 77 & 91.07 / 80.18 \\ \hline
        \texttt{en}$\rightarrow$\texttt{pl}     & -                     & -                      & 83.93 / 74.21 & 74.44 / 62.37 & 70.87 / 59.21  & 81.29 / 69.89 & 84.64 / 74.18 \\ \hline
        \texttt{en}$\rightarrow$\texttt{pt}     & 86.16 / 76.24         & 87.98 / 77.31          & 86.72 / 77.18 & 81.39 / 64.57 & 80.78 / 67.59  & 86.73 / 75.58 & 88.13 / 77.46 \\ \hline
        \texttt{en}$\rightarrow$\texttt{ru}     & 80.39 / 72.45         & 82.6 / 75.19           & 83.09 / 75.77 & 74.54 / 65.96 & 73.71 / 65.9   & 79.43 / 71.83 & 82.54 / 75.36 \\ \hline
        \texttt{en}$\rightarrow$\texttt{th}     & -                     & -                      & 76.73 / 71.07 & 72.5 / 67.44  & 67.83 / 62.15  & 75.73 / 71.41 & 77.55 / 73.14 \\ \hline
        \texttt{en}$\rightarrow$\texttt{tl}     & -                     & -                      & -             & 70.4 / 49.57  & 62.96 / 53.84  & 67.74 / 64.78 & 67.41 / 64.79 \\ \hline
        \texttt{en}$\rightarrow$\texttt{tr}     & -                     & -                      & 76.26 / 73.11 & 66.91 / 63.41 & 63.17 / 58.22  & 75.41 / 72.18 & 77.8 / 75.51  \\ \hline
        \texttt{en}$\rightarrow$\texttt{ur}     & -                     & -                      & 71.94 / 70.89 & 59.45 / 57.74 & 52.63 / 54.42  & 68.93 / 71.17 & 73.29 / 73.38 \\ \hline
        \texttt{en}$\rightarrow$\texttt{vi}     & -                     & -                      & 82.48 / 78.01 & 71.57 / 62.53 & 72.89 / 66.38  & 81.39 / 76.66 & 82.03 / 76.85 \\ \hline
        \texttt{en}$\rightarrow$\texttt{zh}     & 76.74 / 71.62         & 79.42 / 74.13          & 79.82 / 73.73 & 71.67 / 65.01 & 73.35 / 64.68  & 79.84 / 73.73 & 80.85 / 75.29 \\ 
        \end{tabular}
    }
    \caption{Translation performance (XCOMET / COMETKiwi) of different models on the WMT-24 benchmark.}
    \label{tab:wmt24_ours}
    \end{table*}

    \begin{table*}[h]
    \centering
    \resizebox{1\textwidth}{!}{
        \begin{tabular}{c | c c c c c c c}
        Direction & TowerInstruct-7B & TowerInstruct-13B & X-ALMA & Aya-101 & LLaMAX3-Alpaca-8B & GemmaX2-28-2B & GemmaX2-28-9B \\
        \hline 
        \hline
        \texttt{ar}$\rightarrow$\texttt{en} & -                     & -                      & 43.71 / 87.8  & 35.23 / 85.7  & 37.55 / 86.37   & 45.03 / 87.6 & 49.35 / 88.61  \\ 
        \texttt{ar}$\leftarrow$\texttt{en}  & -                     & -                      & 40.73 / 87.74 & 25.07 / 82.98 & 27.14 / 82.49   & 38.52 / 86.37  & 42.9 /  87.76  \\ \hline
        \texttt{bn}$\rightarrow$\texttt{en} & -                     & -                      & -             & 30.06 / 86.45 & 32.52 / 87.65   & 39.29 / 88.74 & 42.76 / 89.71  \\ 
        \texttt{bn}$\leftarrow$\texttt{en}  & -                     & -                      & -             & 20.03 / 82.27 & 23.08 / 81.3    & 32.9 / 86.21 & 34.83 /  86.59 \\ \hline
        \texttt{cs}$\rightarrow$\texttt{en} & -                     & -                      & 46.05 / 89.06 & 38.22 / 87.37 & 41 / 87.91      & 45.25 / 88.76 & 47.18 / 89.2   \\ 
        \texttt{cs}$\leftarrow$\texttt{en}  & -                     & -                      & 41.93 / 91.71 & 31.28 / 88.51 & 32.18 / 87.39   & 38.72 / 90.62 & 42.69 /  91.64 \\ \hline
        \texttt{de}$\rightarrow$\texttt{en} & 49.18 / 89.49         & 50.02 / 89.63          & 49.7 / 89.73  & 42.01 / 88.13 & 44.21 / 88.6    & 49.04 / 89.43 & 49.89 / 89.61  \\ 
        \texttt{de}$\leftarrow$\texttt{en}  & 43.68 / 87.82         & 45.4 / 88.16           & 45.95 / 88.45 & 34.74 / 85.17 & 36.33 / 85.07   & 44.85 / 87.91 & 47.13 /  88.5  \\ \hline
        \texttt{es}$\rightarrow$\texttt{en} & 37.67 / 87.46         & 37.56 / 87.68          & 38.1 / 87.89  & 32.47 / 86.34 & 33.91 / 86.79   & 36.07 / 87.33 & 37.88 / 87.71  \\ 
        \texttt{es}$\leftarrow$\texttt{en}  & 33.09 / 86.79         & 33.98 / 87.05          & 33.67 / 87    & 28.2 / 84.96  & 28.58 / 85.13   & 34.34 / 86.81 & 35.45 /  87.23 \\ \hline
        \texttt{fa}$\rightarrow$\texttt{en} & -                     & -                      & 41.12 / 88.5  & 33.13 / 86.67 & 35.27 / 87.32   & 41.67 / 88.42 & 44.95 / 89.18  \\ 
        \texttt{fa}$\leftarrow$\texttt{en}  & -                     & -                      & 35.46 / 88.35 & 26.39 / 84.98 & 26.63 / 84.18   & 35.49 / 88 & 38.66 /  88.88 \\ \hline
        \texttt{fr}$\rightarrow$\texttt{en} & 50.98 / 89.55         & 51.32 / 89.68          & 49.89 / 89.56 & 43.78 / 88.27 & 45.2 / 88.47    & 50.62 / 89.48 & 51.18 / 89.64  \\ 
        \texttt{fr}$\leftarrow$\texttt{en}  & 53.85 / 88.42         & 55.65 / 88.94          & 55.39 / 88.73 & 43.15 / 85.26 & 45 / 85.79      & 54.2 / 88.39 & 57.69 /  89.11 \\ \hline
        \texttt{he}$\rightarrow$\texttt{en} & -                     & -                      & 46.34 / 88.6  & 38.56 / 86.67 & 41.22 / 87.23   & 48.59 / 88.86 & 51.58 / 89.37  \\ 
        \texttt{he}$\leftarrow$\texttt{en}  & -                     & -                      & 44.45 / 89.26 & 26.74 / 83.65 & 31.36 / 83.35   & 41.96 / 88.48 & 46.26 /  89.26 \\ \hline
        \texttt{hi}$\rightarrow$\texttt{en} & -                     & -                      & 44.63 / 89.81 & 35.65 / 87.84 & 37.05 / 88.51   & 45.71 / 89.88 & 49.14 / 90.53  \\ 
        \texttt{hi}$\leftarrow$\texttt{en}  & -                     & -                      & 38.46 / 81.45 & 24.29 / 76.16 & 27.68 / 75.99   & 37.86 / 80.17 & 41.27 /  81.1  \\ \hline
        \texttt{id}$\rightarrow$\texttt{en} & -                     & -                      & 48.7 / 89.85  & 40.92 / 88.37 & 43.07 / 88.83   & 51.21 / 89.95 & 53.27 / 90.37  \\ 
        \texttt{id}$\leftarrow$\texttt{en}  & -                     & -                      & 49.58 / 92.07 & 38.06 / 89.85 & 38.62 / 89.18   & 51.01 / 91.88 & 52.61 /  92.24 \\ \hline
        \texttt{it}$\rightarrow$\texttt{en} & 40.54 / 88.4          & 41.17 / 88.46          & 40.61 / 88.46 & 35.28 / 87.12 & 35.77 / 87.39   & 39.84 / 88.24 & 42.06 / 88.59  \\ 
        \texttt{it}$\leftarrow$\texttt{en}  & 36.5 / 88.79          & 38.17 / 89.18          & 37.72 / 89.16 & 29.4 / 86.25  & 31.09 / 86.46   & 36.98 / 88.72 & 38.83 /  89.25 \\ \hline
        \texttt{ja}$\rightarrow$\texttt{en} & -                     & -                      & 32.5 / 88.22  & 23.99 / 85.6  & 27.73 / 87.18   & 33.43 / 88.19 & 36.5 / 88.78   \\ 
        \texttt{ja}$\leftarrow$\texttt{en}  & -                     & -                      & 29.58 / 91.19 & 21 / 88.79    & 22.01 / 88.94   & 30.67 / 90.94 & 33.26 /  91.29 \\ \hline
        \texttt{km}$\rightarrow$\texttt{en} & -                     & -                      & -             & 28.08 / 83.98 & 28.74 / 85.69   & 37 / 87.56 & 41.83 / 88.56  \\ 
        \texttt{km}$\leftarrow$\texttt{en}  & -                     & -                      & -             & 19.48 / 81.34 & 14.41 / 72.92   & 20.73 / 82.32 & 24.13 /  84.22 \\ \hline
        \texttt{ko}$\rightarrow$\texttt{en} & 34.51 / 88.19         & 36.01 / 88.51          & 34.31 / 88.31 & 27.38 / 86.53 & 29.07 / 87.13   & 35.35 / 88.44 & 38.76 / 89.04  \\ 
        \texttt{ko}$\leftarrow$\texttt{en}  & 26.39 / 89.4          & 28.39 / 89.87          & 25.74 / 89.27 & 16.34 / 85.94 & 19.62 / 86.79   & 26.7 / 89.24 & 30.37 /  90.13 \\ \hline
        \texttt{lo}$\rightarrow$\texttt{en} & -                     & -                      & -             & 33.44 / 85.25 & 25.96 / 83.19   & 39.44 / 87.51 & 44.22 / 88.82  \\ 
        \texttt{lo}$\leftarrow$\texttt{en}  & -                     & -                      & -             & 28.85 / 83.25 & 10.17 / 62.95   & 27.55 / 82.75 & 31.53 /  84.66 \\ \hline
        \texttt{ms}$\rightarrow$\texttt{en} & -                     & -                      & 48.33 / 89.31 & 41.62 / 87.89 & 43.67 / 88.38   & 51.1 / 89.55 & 53.12 / 90.03  \\ 
        \texttt{ms}$\leftarrow$\texttt{en}  & -                     & -                      & 43.23 / 89.9  & 34.21 / 87.11 & 35.7 / 87.39    & 46.09 / 89.82 & 47.03 /  89.89 \\ \hline
        \texttt{my}$\rightarrow$\texttt{en} & -                     & -                      & -             & 20.43 / 80.38 & 22.75 / 84.73   & 29.79 / 86.58 & 36.14 / 88.21  \\ 
        \texttt{my}$\leftarrow$\texttt{en}  & -                     & -                      & -             & 15.24 / 84.99 & 9.9 / 73.82     & 15.7 / 85.69 & 20.14 /  88.15 \\ \hline
        \texttt{nl}$\rightarrow$\texttt{en} & 37.61 / 87.66         & 38.72 / 87.92          & 38.78 / 88.01 & 33.25 / 86.56 & 34.49 / 86.98   & 37.39 / 87.63 & 39.15 / 87.94  \\ 
        \texttt{nl}$\leftarrow$\texttt{en}  & 35.24 / 88.31         & 36.41 / 88.66          & 35.97 / 88.59 & 28.72 / 86.09 & 29.51 / 85.66   & 35.01 / 88.24 & 37.54 /  88.82 \\ \hline
        \texttt{pl}$\rightarrow$\texttt{en} & -                     & -                      & 37.07 / 86.79 & 29.93 / 84.96 & 32.51 / 85.43   & 36.39 / 86.44 & 38.49 / 86.91  \\ 
        \texttt{pl}$\leftarrow$\texttt{en}  & -                     & -                      & 31.89 / 89.79 & 24.96 / 86.52 & 25.35 / 85.92   & 31.56 / 89.16 & 33.5 /  89.88  \\ \hline
        \texttt{pt}$\rightarrow$\texttt{en} & 55.26 / 89.91         & 56.17 / 90.09          & 53.48 / 89.93 & 47.06 / 88.58 & 49.4 / 88.93    & 54.95 / 89.83 & 57.09 / 90.16  \\ 
        \texttt{pt}$\leftarrow$\texttt{en}  & 49.19 / 89.42         & 50.76 / 89.73          & 53.42 / 90.1  & 43.34 / 87.69 & 45.28 / 87.88   & 51.22 / 89.67 & 53.17 /  90.04 \\ \hline
        \texttt{ru}$\rightarrow$\texttt{en} & 40.74 / 86.93         & 42.76 / 87.23          & 40.69 / 87.11 & 33.68 / 85.43 & 36.32 / 86.13   & 40.63 / 86.8 & 43.17 / 87.28  \\ 
        \texttt{ru}$\leftarrow$\texttt{en}  & 38.37 / 89.4          & 40.27 / 89.93          & 39.9 / 89.96  & 31.3 / 86.85  & 33.42 / 87.23   & 38.64 / 89.14 & 41.4 /  90.13  \\ \hline
        \texttt{th}$\rightarrow$\texttt{en} & -                     & -                      & 34.65 / 88.24 & 27.69 / 86.2  & 30.73 / 87.33   & 37.19 / 88.59 & 40.33 / 89.25  \\ 
        \texttt{th}$\leftarrow$\texttt{en}  & -                     & -                      & 35.87 / 87.47 & 32.45 / 86.12 & 30.79 / 84.72   & 39.45 / 87.91 & 42.53 /  88.7  \\ \hline
        \texttt{tl}$\rightarrow$\texttt{en} & -                     & -                      & -             & 37.04 / 84.2  & 43.35 / 86.34   & 52.91 / 88.18 & 55.97 / 88.92  \\ 
        \texttt{tl}$\leftarrow$\texttt{en}  & -                     & -                      & -             & 23.43 / 80.55 & 27.08 / 81.19   & 37.17 / 84.46 & 38.15 /  84.52 \\ \hline
        \texttt{tr}$\rightarrow$\texttt{en} & -                     & -                      & 43.63 / 89.73 & 34.67 / 87.76 & 36.62 / 88.31   & 44.43 / 89.72 & 47.97 / 90.47  \\ 
        \texttt{tr}$\leftarrow$\texttt{en}  & -                     & -                      & 37.61 / 89.9  & 28.61 / 86.93 & 24.55 / 85.32   & 39.25 / 89.74 & 42.17 /  90.52 \\ \hline
        \texttt{ur}$\rightarrow$\texttt{en} & -                     & -                      & 38.35 / 87.59 & 30.23 / 84.98 & 32.21 / 86.08   & 38.91 / 87.62 & 43.15 / 88.61  \\ 
        \texttt{ur}$\leftarrow$\texttt{en}  & -                     & -                      & 29.56 / 83.38 & 17.78 / 77.56 & 20 / 75.32      & 28.43 / 82.54 & 30.89 /  83.93 \\ \hline
        \texttt{vi}$\rightarrow$\texttt{en} & -                     & -                      & 41.5 / 88.03  & 35.02 / 86.09 & 36.71 / 86.95   & 42.73 / 87.98 & 45.45 / 88.53  \\ 
        \texttt{vi}$\leftarrow$\texttt{en}  & -                     & -                      & 44.21 / 89.64 & 30.71 / 84.93 & 36.53 / 86.59   & 44.67 / 89.4 & 46.65 /  89.96 \\ \hline
        \texttt{zh}$\rightarrow$\texttt{en} & 33.27 / 86.99         & 34.56 / 87.32          & 33.94 / 87.32 & 24.85 / 84.38 & 29.9 / 86.4     & 34.35 / 87.29 & 36.37 / 87.62  \\ 
        \texttt{zh}$\leftarrow$\texttt{en}  & 33.9 / 87.77          & 36.35 / 88.45          & 34.43 / 88.04 & 24.24 / 84.81 & 26.96 / 85.4    & 39.39 / 88.88 & 41.6 /  89.18  \\ \hline
        \end{tabular}
    }
    \caption{English-centric translation performance (spBLEU / COMET) of different models on the FLORES-200 benchmark.}
    \label{tab:flores200_en_ours}
    \end{table*}

    \begin{table*}[h]
    \centering
    \resizebox{1\textwidth}{!}{
        \begin{tabular}{c | c c c c c c}
        Direction & TowerInstruct-7B & TowerInstruct-13B & Aya-101 & LLaMAX3-Alpaca-8B & GemmaX2-28-2B & GemmaX2-28-9B \\
        \hline
        \hline
        \texttt{ar}$\rightarrow$\texttt{zh} & -                     & -                      & 19.62 / 81.35 & 21.34 / 82.33    & 30.33 / 85.23  & 34.11 / 86.37 \\ 
        \texttt{ar}$\leftarrow$\texttt{zh}  & -                     & -                      & 16.31 / 79.75 & 18.35 / 80.54    & 23.2 / 82.56  & 27.62 / 84.39 \\ \hline
        \texttt{bn}$\rightarrow$\texttt{zh}    & -                     & -                      & 15.91 / 79.83 & 19.3 / 83.08  & 27.13 / 85.48   & 31.43 / 87.14 \\ 
        \texttt{bn}$\leftarrow$\texttt{zh}     & -                     & -                      & 11.43 / 76.82 & 14.78 / 77.21  & 20.71 / 81.4  & 24.57 / 82.68 \\ \hline
        \texttt{cs}$\rightarrow$\texttt{zh}     & -                     & -                      & 19.73 / 82.54 & 22.82 / 83.86 & 31.8 / 86.71   & 35.03 / 87.46 \\ 
        \texttt{cs}$\leftarrow$\texttt{zh}     & -                     & -                      & 17.93 / 85.82 & 18.86 / 86.4   & 23.55 / 87.96  & 28.17 / 89.81 \\ \hline
        \texttt{de}$\rightarrow$\texttt{zh}     & 28.43 / 85.96         & 30.43 / 86.71          & 21.55 / 83.32 & 23.39 / 84.54 & 33.77 / 87.24   & 35.49 / 87.78 \\ 
        \texttt{de}$\leftarrow$\texttt{zh}     & 23.93 / 83.48         & 26.53 / 84.66          & 18.52 / 81.28 & 21.08 / 82.65  & 26.56 / 84.19  & 29.29 / 85.39 \\ \hline
        \texttt{es}$\rightarrow$\texttt{zh}     & 25.82 / 85.83         & 27.93 / 86.68          & 24.24 / 84.81 & 26.96 / 85.4  & 30.93 / 87.33  & 32.84 / 87.9  \\ 
        \texttt{es}$\leftarrow$\texttt{zh}     & 21.34 / 84.21         & 22.84 / 84.76          & 24.85 / 84.38 & 29.9 / 86.4    & 23.51 / 84.77  & 25.45 / 85.54 \\ \hline
        \texttt{en}$\rightarrow$\texttt{zh}     & 33.9 / 87.77          & 36.35 / 88.45          & 19.96 / 83.76 & 21.3 / 84.33  & 39.39 / 88.87   & 41.64 / 89.19 \\ 
        \texttt{en}$\leftarrow$\texttt{zh}     & 33.27 / 86.99         & 34.56 / 87.32          & 17.63 / 82.42 & 19.42 / 83.81  & 34.23 / 87.3  & 36.37 / 87.63 \\ \hline
        \texttt{fa}$\rightarrow$\texttt{zh}     & -                     & -                      & 18.66 / 82.13 & 20.83 / 83.74 & 29.39 / 86.16   & 33.12 / 87.17 \\ 
        \texttt{fa}$\leftarrow$\texttt{zh}     & -                     & -                      & 15.92 / 81.43 & 17.85 / 82.72  & 23.56 / 85.04  & 26.12 / 86.01 \\ \hline
        \texttt{fr}$\rightarrow$\texttt{zh}     & 28.84 / 86.2          & 31.18 / 86.94          & 21.77 / 83.83 & 23.81 / 84.59 & 33.59 / 87.34   & 35.68 / 87.94 \\ 
        \texttt{fr}$\leftarrow$\texttt{zh}     & 30.07 / 83.92         & 31.55 / 84.96          & 24.29 / 80.98 & 26.03 / 83.06  & 31.93 / 84.77  & 35.25 / 85.63 \\ \hline
        \texttt{he}$\rightarrow$\texttt{zh}     & -                     & -                      & 19.48 / 81.98 & 21.91 / 82.77 & 31.13 / 85.78   & 35.14 / 86.95 \\ 
        \texttt{he}$\leftarrow$\texttt{zh}     & -                     & -                      & 13.92 / 79.22 & 18.36 / 81.24  & 23.12 / 83.65  & 28.53 / 86.03 \\ \hline
        \texttt{hi}$\rightarrow$\texttt{zh}     & -                     & -                      & 19.77 / 82.8  & 20.59 / 83.04 & 29.74 / 86.05   & 33.27 / 87.18 \\ 
        \texttt{hi}$\leftarrow$\texttt{zh}     & -                     & -                      & 13.86 / 68.98 & 16.89 / 70.61  & 21.54 / 72.62  & 25.41 / 74.49 \\ \hline
        \texttt{id}$\rightarrow$\texttt{zh}     & -                     & -                      & 19.99 / 82.24 & 23.17 / 83.98 & 33.4 / 86.96   & 35.79 / 87.7  \\ 
        \texttt{id}$\leftarrow$\texttt{zh}     & -                     & -                      & 19.72 / 86.1  & 20.77 / 86.26  & 30.08 / 88.61  & 32.34 / 89.21 \\ \hline
        \texttt{it}$\rightarrow$\texttt{zh}     & 26.61 / 86.14         & 29.15 / 86.9           & 19.82 / 83.5  & 22.21 / 84.43 & 31.61 / 87.3   & 33.93 / 87.89 \\ 
        \texttt{it}$\leftarrow$\texttt{zh}     & 22.33 / 86.17         & 23.68 / 86.71          & 17.38 / 83.25 & 19.88 / 84.92  & 24.38 / 86.4  & 27.14 / 87.36 \\ \hline
        \texttt{ja}$\rightarrow$\texttt{zh}     & -                     & -                      & 18.68 / 85.02 & 19.6 / 85.48  & 28.88 / 87.81   & 32.02 / 88.6  \\ 
        \texttt{ja}$\leftarrow$\texttt{zh}     & -                     & -                      & 16.86 / 87.91 & 15.73 / 88.04  & 22.28 / 89.75  & 26.1 / 90.55  \\ \hline
        \texttt{km}$\rightarrow$\texttt{zh}     & -                     & -                      & 7.71 / 69.75  & 18.11 / 81.95 & 26.48 / 85.46   & 30.82 / 86.78 \\ 
        \texttt{km}$\leftarrow$\texttt{zh}     & -                     & -                      & 15.84 / 78.6  & 9.15 / 67.7    & 18.61 / 80.24  & 21.99 / 81.93 \\ \hline
        \texttt{ko}$\rightarrow$\texttt{zh}     & 24.11 / 85.1          & 27.83 / 86.64          & 19.58 / 83.73 & 21.32 / 84.95 & 29.63 / 86.98   & 32.62 / 88    \\ 
        \texttt{ko}$\leftarrow$\texttt{zh}     & 17.69 / 86.5          & 20.11 / 87.41          & 11.92 / 83.3  & 14.37 / 85.51  & 19.75 / 87.06  & 22.77 / 88.13 \\ \hline
        \texttt{lo}$\rightarrow$\texttt{zh}     & -                     & -                      & 17.09 / 80.82 & 13.85 / 79.06 & 26.95 / 85.08   & 31.63 / 86.91 \\ 
        \texttt{lo}$\leftarrow$\texttt{zh}     & -                     & -                      & 18.84 / 80.25 & 6.28 / 58.96   & 20.59 / 80.34  & 25.93 / 82.88 \\ \hline
        \texttt{ms}$\rightarrow$\texttt{zh}     & -                     & -                      & 18.84 / 81.08 & 22.45 / 82.88 & 32.8 / 86.53   & 35.42 / 87.41 \\ 
        \texttt{ms}$\leftarrow$\texttt{zh}     & -                     & -                      & 17.94 / 82.92 & 18.16 / 83.91  & 26.87 / 86.44  & 28.57 / 86.72 \\ \hline
        \texttt{my}$\rightarrow$\texttt{zh}     & -                     & -                      & 9.01 / 75.51  & 12.78 / 80.9  & 21.69 / 83.92   & 27.63 / 86.06 \\ 
        \texttt{my}$\leftarrow$\texttt{zh}     & -                     & -                      & 11.44 / 81.88 & 7.3 / 67.8     & 11.32 / 81.5  & 15.21 / 84.47 \\ \hline
        \texttt{nl}$\rightarrow$\texttt{zh}     & 25.34 / 84.97         & 27.25 / 85.99          & 18.37 / 81.77 & 20.97 / 83.19 & 29.85 / 86.28   & 31.65 / 86.93 \\ 
        \texttt{nl}$\leftarrow$\texttt{zh}     & 22.46 / 85.32         & 23.31 / 85.84          & 17.88 / 82.86 & 19.76 / 84.14  & 23.63 / 85.4  & 26.92 / 86.43 \\ \hline
        \texttt{pl}$\rightarrow$\texttt{zh}     & -                     & -                      & 19.48 / 82.3  & 20.77 / 83.18  & 29.6 / 85.95  & 32.2 / 86.77  \\ 
        \texttt{pl}$\leftarrow$\texttt{zh}     & -                     & -                      & 16.36 / 84.62 & 17.9 / 85.49    & 23.38 / 87.84 & 25.72 / 89.03 \\ \hline
        \texttt{pt}$\rightarrow$\texttt{zh}     & 29.02 / 86.37         & 31.32 / 87.19          & 21.02 / 83.84 & 23.65 / 84.7   & 34.24 / 87.73  & 36.88 / 88.35 \\ 
        \texttt{pt}$\leftarrow$\texttt{zh}     & 26.18 / 85.91         & 28.02 / 86.73          & 22.4 / 83.97  & 25.19 / 85.53   & 30.37 / 86.78 & 33.01 / 87.38 \\ \hline
        \texttt{ru}$\rightarrow$\texttt{zh}     & 26.81 / 84.94         & 29.46 / 85.66          & 20.65 / 82.63 & 22.75 / 83.3   & 31.67 / 86.15  & 33.73 / 86.84 \\ 
        \texttt{ru}$\leftarrow$\texttt{zh}     & 22.88 / 86.18         & 26.17 / 87.99          & 19.18 / 85.34 & 22.02 / 86.89   & 25.58 / 87.42 & 28.57 / 88.82 \\ \hline
        \texttt{th}$\rightarrow$\texttt{zh}     & -                     & -                      & 19.08 / 83.36 & 21.09 / 85.05  & 29.61 / 87.11  & 32.53 / 88.16 \\ 
        \texttt{th}$\leftarrow$\texttt{zh}     & -                     & -                      & 25.67 / 83.94 & 24.4 / 82.97    & 31.55 / 85.87 & 35.18 / 86.75 \\ \hline
        \texttt{tl}$\rightarrow$\texttt{zh}     & -                     & -                      & 18.78 / 78.78 & 21.36 / 81.07  & 32.3 / 85.05  & 35.63 / 86.02 \\ 
        \texttt{tl}$\leftarrow$\texttt{zh}     & -                     & -                      & 13.76 / 77.03 & 15.74 / 77.96   & 22.2 / 81 & 24.19 / 81.87 \\ \hline
        \texttt{tr}$\rightarrow$\texttt{zh}     & -                     & -                      & 19.02 / 81.57 & 21.59 / 83.16  & 30.97 / 86.17  & 34.79 / 87.29 \\ 
        \texttt{tr}$\leftarrow$\texttt{zh}     & -                     & -                      & 16.5 / 81.75  & 13.94 / 81.37   & 23.47 / 84.77 & 26.43 / 86.05 \\ \hline
        \texttt{ur}$\rightarrow$\texttt{zh}     & -                     & -                      & 16.77 / 79.86 & 19.15 / 81.54  & 26.85 / 84.84  & 31.67 / 86.48 \\ 
        \texttt{ur}$\leftarrow$\texttt{zh}     & -                     & -                      & 10.98 / 73.4  & 13.57 / 73.03   & 17.71 / 77.73 & 21.42 / 79.73 \\ \hline
        \texttt{vi}$\rightarrow$\texttt{zh}     & -                     & -                      & 20.43 / 84.24 & 22 / 84.89     & 32.15 / 87.43  & 34.26 / 87.95 \\ 
        \texttt{vi}$\leftarrow$\texttt{zh}     & -                     & -                      & 21.31 / 83.98 & 24.81 / 85.98   & 31.67 / 88.35 & 33.79 / 88.8  \\ \hline
        \end{tabular}
    }
    \caption{Chinese-centric translation performance (spBLEU / COMET) of different models on the FLORES-200 benchmark.}
    \label{tab:flores200_zh_ours}
    \end{table*}

\end{document}